%% file: main_arxiv.tex
\documentclass[runningheads]{llncs}

\input{paper_info}

\usepackage{eccv}

\usepackage{eccvabbrv}

\usepackage{graphicx}
\usepackage{booktabs}

\input{preamble}

\usepackage{multirow}
\usepackage{enumitem}
\usepackage{makecell}
\usepackage{amsmath}
\usepackage{amssymb}
\usepackage{graphicx}
\usepackage{subcaption}
\usepackage{gensymb}
\usepackage{gensymb}
\usepackage{adjustbox}
\usepackage{wrapfig}
\usepackage[final]{pdfpages}

\input{macros}

\usepackage[accsupp]{axessibility}  

\usepackage{hyperref}

\usepackage{orcidlink}

\begin{document}

\title{\ourPaperTitle}


\author{
In Cho\orcidlink{0009-0006-2131-4430} \and
Hyunbo Shim\orcidlink{0009-0001-6408-1992} \and
Seon Joo Kim\orcidlink{0000-0001-8512-216X}
}

\authorrunning{I. Cho et al.}

\institute{Yonsei University}

\maketitle

\input{Main/Sections/abstract}
\input{Main/Sections/intro}
\input{Main/Sections/related_work}
\input{Main/Sections/method}
\input{Main/Sections/experiment}
\input{Main/Sections/discussion}
\input{Main/Sections/conclusion}


%
%
\bibliographystyle{splncs04}
\bibliography{main}

\clearpage
\title{(Supplementary Material)\\ \ourPaperTitle}
\authorrunning{I. Cho et al.}
\titlerunning{(Supplementary) Learning to Enhance Aperture Phasor Field}
\author{}
\institute{}
\maketitle

\setcounter{figure}{8}
\setcounter{table}{3}
\setcounter{equation}{7}
\setcounter{page}{19}

\input{Supple/supple}
\clearpage

\end{document}

%% file: paper_info.tex
\def\ourPaperTitle {Learning to Enhance Aperture Phasor Field\\ for Non-Line-of-Sight Imaging}

%% file: preamble.tex
\usepackage[dvipsnames]{xcolor}


%% file: macros.tex
\newcommand{\Tref}[1]{Table~\ref{#1}}

\newcommand{\Fref}[1]{Fig.~\ref{#1}}

\newcommand{\Sref}[1]{Section~\ref{#1}}

\newcommand{\textblock}[1]{\noindent\textbf{#1}}
\newcommand{\sqsize}[4]{$#1\ \textrm{#2} \times #3\ \textrm{#4}$}

\newcommand{\parabf}[1]{\paragraph{\textbf{\upshape #1}}}

\def\naive{na\"ive\ }
\def\eg{\emph{e.g}\onedot} 

\def\ie{\emph{i.e}\onedot}

\def\etal{\emph{et al}\onedot}


%% file: Main/Sections/abstract.tex
\begin{abstract}
This paper aims to facilitate more practical NLOS imaging by reducing the number of samplings and scan areas.
To this end, we introduce a phasor-based enhancement network that is capable of predicting clean and full measurements from noisy partial observations.
We leverage a denoising autoencoder scheme to acquire rich and noise-robust representations in the measurement space.
Through this pipeline, our enhancement network is trained to accurately reconstruct complete measurements from their corrupted and partial counterparts.
However, we observe that the \naive application of denoising often yields degraded and over-smoothed results, caused by unnecessary and spurious frequency signals present in measurements.
To address this issue, we introduce a phasor-based pipeline designed to limit the spectrum of our network to the frequency range of interests, where the majority of informative signals are detected.
The phasor wavefronts at the aperture, which are band-limited signals, are employed as inputs and outputs of the network, guiding our network to learn from the frequency range of interests and discard unnecessary information.
The experimental results in more practical acquisition scenarios demonstrate that we can look around the corners with $16\times$ or $64\times$ fewer samplings and $4\times$ smaller apertures.
Our code is available at \url{https://github.com/join16/LEAP}.

\keywords{Non-line-of-sight imaging \and Deep learning}
\end{abstract}

%% file: Main/Sections/intro.tex
\section{Introduction}
Non-line-of-sight (NLOS) imaging aims to reconstruct scenes that are hidden in direct line-of-sight systems, with a laser illuminating a relay wall, and a time-resolved detector recording the returning photons.
The ability to perceive occluded objects has captivated many researchers due to its wide range of future applications, such as medical imaging, rescue operations, and autonomous driving.
Representative NLOS imaging methods \cite{o2018lct, young2020dlct, lindell2019fk, liu2019phasor, liu2020diffraction} have shown that the hidden scenes can be reconstructed in high-quality if measurements are captured with sufficient sampling points, acquisition time, and scanning areas.

\input{Main/Partials/figure_intro}
Beyond the recent advances, we extend our focus to more practical acquisition scenarios relevant to real-world applications, most of which do not offer sufficient scanning time and areas.
Reducing the scan areas, the number of samplings, and thereby the scanning time yields much degraded results in previous methods, due to their theoretical resolution limits and increased effects of noise.
To mitigate this issue, recent studies leverage custom-designed arrays of single-photon avalanche diode (SPAD) sensors \cite{nam2021low, pei2021dynamic, mu2022rescue} or optimization-based methods \cite{ye2021compressed, liu2023sscr}, incurring additional expenses for hardware or computations. 

In this paper, we introduce a phasor-based enhancement network that leverages learned priors and phasor-based frequency filtering to facilitate NLOS imaging under more practical acquisition setups (\Fref{fig:intro} (b)), where measurements are acquired with fewer samplings, smaller scan areas, and reduced acquisition time.
We begin by tailoring a denoising autoencoder scheme \cite{vincent2008dae}, where we place the denoising criterion on top of the missing signal recovery problem \cite{wang2023ssn}. This enables our model to attain rich and noise robust representations in the measurement space.
Specifically, our network processes partial measurements corrupted by Poisson noise \cite{hernandez2017spad, o2017transient, chen2020lfe}, and is trained to accurately predict the optimal measurements, containing sufficient scanning points and clean signals.
After training, our method enables high-quality NLOS reconstruction in partial sampling scenarios, which is achieved by applying inverse NLOS methods to the predictions.

While this straightforward application of the denoising criterion alleviates the effects of noise, this training scheme leads the enhancement network to be parameterized across the entire frequency spectrum, often producing degraded and over-smoothed results.
Conversely, in measurements of NLOS imaging, the majority of informative signals are concentrated within a specific frequency range while signals in other frequency ranges mostly contain coarse structures and noise.
We observe that these unnecessary and spurious frequency signals are key factors causing such degradation, but do not contribute to the reconstruction of hidden volumes.
Thus, we aim to prune them from our network's interests.

To this end, we propose a phasor-based scheme that utilizes the phasor field at the aperture for supervision.
In phasor field NLOS methods \cite{liu2019phasor, liu2020diffraction}, aperture wavefronts are computed by the convolution of the measurements and the illumination function, typically defined as Gaussian in the frequency domain.
The aperture wavefronts are thus signals with a limited frequency band, where the majority of the informative signals are observed.
By leveraging these band-limited signals as inputs and outputs of the network, we constrain the network operations within the frequency range of interests.
This guides our network to discard unnecessary signals, achieving substantial improvement in reconstruction quality and better generalization capability to real-world measurements.

Coupling the phasor-based network with the denoising autoencoding scheme, we name our method as Learning to Enhance Aperture Phasor field (LEAP).
We validate our model in sparse sampling and smaller aperture scenarios, on both confocal and non-confocal measurements.
The experimental results showcase the effectiveness of our phasor-based enhancement network, demonstrating that we can look around the corners with $16\times$ or $64\times$ fewer samplings, and $4\times$ smaller apertures, all without incurring additional costs.

%% file: Main/Partials/figure_intro.tex
\begin{figure}[htb!]
    \centering
    \includegraphics[trim={0 20 0 0pt}, width=1\linewidth]{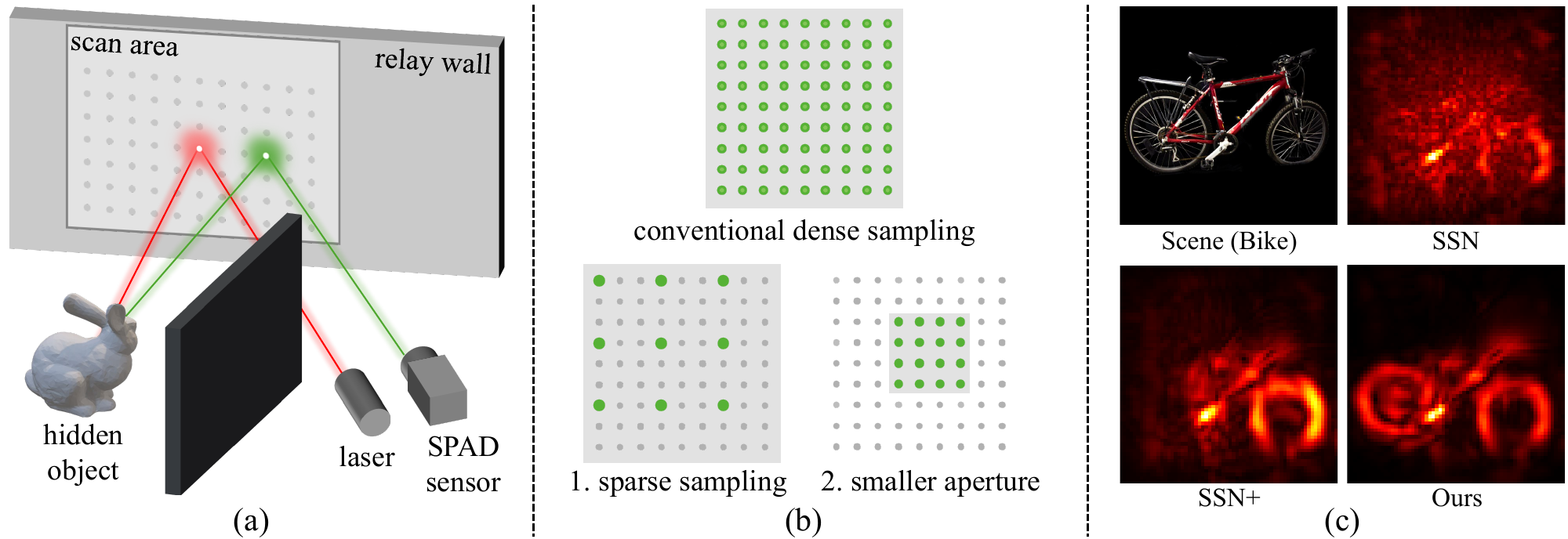}
    \caption{\textbf{(a)} A typical NLOS imaging system. \textbf{(b)} More practical acquisition scenarios of NLOS imaging: sparse sampling and scanning with smaller apertures. 
    \textbf{(c)} Results on confocal $16 \times 16$ measurements of Bike \cite{lindell2019fk}. Our method exhibits high-quality results with $16 \times$ fewer sampling points and a shorter acquisition time, whereas previous signal recovery network (SSN) \cite{wang2023ssn} and simple addition of the denoising criterion to SSN (SSN+) fail to correctly reconstruct the hidden objects.}
    \label{fig:intro}
\end{figure}

%% file: Main/Sections/related_work.tex
\section{Related Work}
\textblock{NLOS imaging and methods.}
The concept of NLOS imaging was originally proposed in \cite{kirmani2009looking} and experimentally validated in \cite{velten2013femto}. It has been further developed in subsequent research involving SPAD sensors.
Methods for NLOS imaging can be broadly divided into two streams: active \cite{o2018lct, lindell2019fk, isogawa2020optical, tsai2017first, isogawa2020efficient} methods with a controllable light source and passive methods relying on the indirect light \cite{seidel2020two, tanaka2020polarized, saunders2019computational, boger2019passive, batarseh2018passive, yedidia2019using}.
We follow the line of the active methods using a SPAD sensor and a laser, which usually offer a wider range of reconstructable objects.

Early methods of active NLOS imaging propose back-projection based solutions \cite{velten2012fbp, laurenzis2014nonline, arellano2017fast, ahn2019convolutional} with $O(N^5)$ computational complexity.
Successive researches \cite{o2018lct, young2020dlct, lindell2019fk, liu2020diffraction} alleviate such computational costs by fast Fourier transform (FFT) based inverse methods with $O(N^3logN)$ complexity.
These include Light Cone Transform (LCT) \cite{o2018lct} based on 3D convolution, DLCT with a vector deconvolution \cite{young2020dlct}, and the wave-based solution \cite{lindell2019fk} using Stolt's method.

Recently proposed phasor field NLOS methods \cite{liu2019phasor, liu2020diffraction} formulate NLOS imaging as a wave propagation problem.
These methods show that NLOS imaging can be solved with well-established line-of-sight propagation operators such as the Rayleigh-Sommerfeld diffraction (RSD) integral.
The faster RSD algorithm with ring and radius-based samplings \cite{jiang2021ring} further boosts up the efficiency.
These methods, including the phasor field and other inverse methods, exhibit remarkable results on measurements captured with sufficient samplings and scan areas.
Our goal is to extend these recent advances to more practical scanning setups.

\textblock{NLOS imaging in practical scenarios.}
Several attempts \cite{nam2021low, pei2021dynamic, ye2021compressed, liu2023sscr, isogawa2020c2nlos, willomitzer2021holo, liao2021fpga} have been made to address NLOS imaging with reduced acquisition time.
One promising approach is increasing the number of pixels with the arrays of high-end SPAD sensors \cite{nam2021low, pei2021dynamic}.
Despite clear advantages, SPAD arrays also introduce additional hardware costs depending on the number of sensing pixels.
Another stream of works explores optimization-based algorithms for sparse sampling scenarios \cite{ye2021compressed, liu2023sscr}, which suffer from huge computational costs of the optimization.
Apart from these works, our method achieves more practical NLOS imaging with negligible computational costs and without additional hardware.

\textblock{Deep learning and NLOS.}
A number of recent methods employ neural networks for NLOS imaging \cite{chopite2020deep, chen2020lfe, shen2021netf, mu2022rescue, plack2023rendering, li2023nlost, zhu2022remapping}.
Due to the generalization issue of the first learning-based method \cite{chopite2020deep}, successive studies \cite{chen2020lfe, mu2022rescue, li2023nlost, yu2023lik} employ the physics-based models after the lightweight convolution layers, and focus on refining propagated spatial feature volumes with learned priors.
These models rely on the employed propagators and have insufficient capacity to extract meaningful representations in partial sampling scenarios.
Recently, to address NLOS imaging with fewer samplings, Signal Super-resolution Network (SSN) \cite{wang2023ssn} performs super-resolution on sparsely sampled measurements, and Li \etal extend LFE \cite{chen2020lfe} by employing signal recovery network before propagating feature volumes.
However, solitary learning of the signal recovery problem leads SSN to be vulnerable to noise.
We also illustrate that the \naive application of the denoising criterion or its incorporation with volume refinement \cite{li2024usm} often fail to bring meaningful improvement, highlighting the necessity of adequate frequency management.

%% file: Main/Sections/method.tex
\section{Proposed Method}
\subsection{Preliminary of Phasor Field NLOS Imaging}
The goal of NLOS imaging is to reconstruct hidden scenes from measurements of indirect multi-bounce light reflections. 
Short laser pulses illuminate a set of points $\mathbf{x_p}$ on a relay wall $P$. The light scatters towards the hidden object and some photons hit the object and return back to the relay wall. Scanning a set of points $\mathbf{x_c}$ on a relay wall $C$ produces the impulse response $H(\mathbf{x_p \rightarrow x_c}, t)$.

Recent phasor field NLOS methods have demonstrated that NLOS imaging can be viewed as a diffractive wave propagation problem with a virtual camera, which can be solved with line-of-sight diffraction operators \cite{liu2019phasor, liu2020diffraction}. The phasor wavefront at the virtual aperture can be computed from $H(\mathbf{x_p \rightarrow x_c}, t)$:
\begin{equation}
    \mathcal{P}(\mathbf{x_c}, t) = \int_{P}{[\mathcal{P}(\mathbf{x_p}, t)\ *\ H(\mathbf{x_p \rightarrow x_c}, t)]\ \mathbf{dx_p}},
\end{equation}
where $\mathcal{P}(\mathbf{x_p}, t)$ is the wavefront of the virtual illumination source and $*$ is the convolution operator in time.
The hidden scenes can be reconstructed from $\mathcal{P}(\mathbf{x_c}, t)$ using the wave propagation operator $\Phi(\cdot)$:
\begin{equation}
    I(\mathbf{x_v}) = \Phi(\mathcal{P}(\mathbf{x_c}, t)),
\end{equation}
where $\mathbf{x_v}$ is a point in the hidden scenes being imaged.
The propagation operator $\Phi(\cdot)$ is commonly formulated using the Rayleigh-Sommerfeld Diffraction (RSD) integral \cite{liu2020diffraction, jiang2021ring}.
Despite their remarkable results, the reconstruction quality of diffraction-based NLOS methods depends on the quality of the measurements.

\textblock{Resolution limit.}
The spatial resolution of the phasor camera is determined as $0.61\lambda L/d$, where $\lambda$ is the wavelength, $L$ is the imaging distance and $d$ is the diameter of the virtual aperture \cite{liu2019phasor}.
Since the minimum achievable wavelength is determined by the sampling distance $\Delta_p$ ($\lambda > 2\Delta_p$), increasing $\Delta_p$ or reducing $d$ theoretically limits the spatial resolution of the systems.
We aim to increase the achievable resolution of the imaging systems by exploiting learned priors, to recover full measurements from partial inputs.

\textblock{Illumination phasor field.}
The phasor field NLOS imaging is mostly implemented as a virtual transient camera with a short Gaussian shape flash \cite{liu2019phasor, liu2020diffraction}.
The corresponding illumination function is $\mathcal{P}(\mathbf{x_p}, t) = \delta(\mathbf{x_p}-\mathbf{x_{ls}})(e^{i \Omega_C t}e^{-\frac{t^2}{2 \sigma^2}})$, and its Fourier domain representation can be expressed as
\begin{equation}
\mathcal{P}_{\mathcal{F}}(\mathbf{x_p}, \Omega) =
\delta(\mathbf{x_p}-\mathbf{x_{ls}})\ (2\pi\delta(\Omega - \Omega_C)\ *\ \sigma\sqrt{2\pi}e^{-\frac{\sigma^2 \Omega^2}{2}}),
\end{equation}
where $\mathbf{x_{ls}}$ is the virtual light source position and $\Omega_C$ is the central frequency determined by the wavelength $\lambda$.
The illumination phasor field in the frequency domain $\mathcal{P}_{\mathcal{F}}(\mathbf{x_p}, \Omega)$ is defined as Gaussian (\Fref{fig:method_frequency}, top-left), which works as a band-pass filter.
The computed phasor wavefront at the aperture is thus band-limited signals, indicating that signals only in a certain frequency range are necessary for reconstructing hidden scenes.

\subsection{Denoising and Frequency of Interests}
NLOS imaging suffers from measurements with an extremely low signal-to-noise ratio (SNR).
Reducing the number of scan points, and thereby reducing the number of total detected photons, amplifies the effects of noise.
Since sensor noise is commonly modeled as Poisson distribution \cite{o2017transient, hernandez2017spad}, we propose to apply the denoising criterion for Poisson noise on top of the signal recovery problem \cite{wang2023ssn}.
Unfortunately, such a training scheme guides the network to recover the entire frequency components, due to the effects of the sensor noise across the entire spectrum \cite{hernandez2017spad}.
The network trained with this scheme often yields degraded and over-smoothed results with missing fine details.

\input{Main/Partials/figure_method_frequency}
To examine the effects of noise on frequency components, we visualize the frequency components of the rendered measurements of Stanford Bunny, both with and without noise.
For better understanding, we also visualize reconstructed scenes using FK \cite{lindell2019fk} and a band-pass filter, retaining frequency components within a given range and discards others (details in Supplement A).
As shown in \Fref{fig:method_frequency}, informative signals are mostly observed around the central frequency of the illumination function (range B).
By adding Poisson noise, some artifacts appear in a lower frequency range and higher frequency components become indistinguishable from noise.
On the other hand, signals near the central frequency still contain clearly visible shapes of objects, are more robust to the noise, and thus are easier to recover.
This motivates us to restrict our network's spectrum, by utilizing the aperture phasor field as band-limited inputs and outputs.

\subsection{Learning to Enhance Aperture Phasor Field}
\label{sec:leap}
Based on the above observations, we propose the phasor-based neural network, coined as Learning to Enhance Aperture Phasor field (LEAP), which can predict clean and full measurements from noisy partial observations.
We assume a single virtual illumination point $\mathbf{x_{ls}}$ for simplicity, which removes the integral over $P$.
    
\Fref{fig:method} depicts the overview of our proposed method.
We begin by sampling partial inputs from full measurements $H(\mathbf{x_p \rightarrow x_c}, t)$ and corrupting them with Poisson noise.
Then the enhancement network takes these noisy partial inputs and predicts the optimal phasor field at the aperture, containing full scans and clean signals, in the frequency domain.
We train our network by minimizing the L1 distance between the predicted and the optimal phasor field at the aperture.
After training, hidden scenes are reconstructed by propagating the predicted phasor field using the RSD algorithm.
We describe details on each part below.

\textblock{Sensor noise simulation.}
To simulate the strong effects of the noise in NLOS imaging, we follow the computational model of SPAD \cite{saunders2019computational, o2017transient, chen2020lfe} and utilize Poisson distribution to model the sensor noise.
Considering the cumulative photon counting procedure, we model the sensor noise with multiple exposure levels as
\begin{equation}
\begin{gathered}
    X = (\eta\tilde{H}(\mathbf{x_p \rightarrow x_c}) * g) + d, \cr
    H'(\mathbf{x_p \rightarrow x_c}, t) \sim Poisson(c \cdot X),
\end{gathered}
\end{equation}
where $H'$ is noised measurements, $\tilde{H}$ is partial measurements subsampled from $H$. 
$\eta$ is the photon detection efficiency and $g$ models the time jitter \cite{o2017transient}.
$c$ controls the exposure time and $d$ models the background noise, including both ambient light and dark counts.
Once the measurements are partially sampled and corrupted with noise, they are taken to the network as inputs.

\input{Main/Partials/figure_method}
\textblock{Input phasor field convolution.}
The noise-augmented partial inputs are then convolved with multiple illumination functions.
We employ a set of illumination wavefronts with multiple wavelengths, of which frequency ranges are chosen to be near the target frequency range.
The convolved outputs $F = \{f_1, f_2, ..., f_i\}$ with the multiple wavelengths $\{\lambda_1, \lambda_2, ..., \lambda_i\}$ are computed in the frequency domain using the convolution theorem, which can be described as
\begin{equation}
    f_i(\mathbf{x_c}, t) = 
    \mathcal{F}^{-1}(
        \mathcal{F}(H'(\mathbf{x_p \rightarrow x_c}), t) 
        \cdot 
        \mathcal{P}^{i}_\mathcal{F}(\mathbf{x_p}, \Omega)
    ),
\end{equation}
where $\mathcal{P}^{i}_\mathcal{F}(\mathbf{x_p}, \Omega)$ is the illumination phasor field in the frequency domain with a wavelength $\lambda_i$.
Both real and imaginary components of $F$ are concatenated and passed to the enhancement network for feature extraction.

\textblock{Enhancement network.}
We employ a 3D residual convolutional neural network (CNN) as the enhancement network. Our network extracts feature volumes from $F$ with several 3D residual blocks, and then transforms them into the frequency domain.
Then 3 convolution layers further extract features from these frequency volumes and predict the residuals, both real and imaginary parts in the frequency domain.
The residuals are then added with the upsampled (and zero-padded in smaller aperture cases) inputs to predict the clean and full measurements $\hat{H}(\mathbf{x_p \rightarrow x_c}, \Omega)$.
Our model finally computes the aperture phasor field:
\begin{equation}
    \hat{\mathcal{P}}_{\mathcal{F}}(\mathbf{x_c}, \Omega) =
    \hat{H}(\mathbf{x_p \rightarrow x_c}, \Omega) \cdot \mathcal{P}_{\mathcal{F}}(\mathbf{x_p}, \Omega),
\end{equation}
where $\mathcal{P}_{\mathcal{F}}(\mathbf{x_p}, \Omega)$ is the target illumination phasor wavefront with the wavelength $\lambda_T$, which will also be used to compute the ground truth phasor field.
We provide details of the network architecture in Supplement B.2.

Compared with SSN \cite{wang2023ssn}, our network only consists of 3D convolution blocks and does not include computationally expensive attention branches, resulting in the improved efficiency of our model (see \Sref{sec:discussion}, runtime analysis).

\textblock{Training objective and reconstruction.}
To train the network, we minimize the L1 distance between the predicted phasor field $\hat{\mathcal{P}}_{\mathcal{F}}(\mathbf{x_c}, \Omega)$ and the target phasor field $\mathcal{P}_{\mathcal{F}}(\mathbf{x_c}, \Omega)$ at the aperture.
The target aperture wavefront is computed by convolving the optimal measurements with the target illumination function $\mathcal{P}_{\mathcal{F}}(\mathbf{x_p}, t)$.
Since the informative frequency components needed for reconstructing hidden scenes are determined by $\lambda_T$, we only minimize the loss for such components.
The training objective of our method can be described as
\begin{equation}
    \mathcal{L} = \sum_{\Omega' \in S}{||\hat{\mathcal{P}}_{\mathcal{F}}(\mathbf{x_c}, \Omega') - \mathcal{P}_{\mathcal{F}}(\mathbf{x_c}, \Omega')||_1},\ \ 
\end{equation}
where $S = [\Omega_C - \Delta\Omega, \Omega_C + \Delta\Omega]$ is a range where the coefficients of input wavefront is larger than a peak ratio $\gamma$, and the central frequency $\Omega_C$ is determined by the target wavelength $\lambda_T$.
Supervising the network with the aperture wavefront restricts the frequency spectrum of the training objective, confining the operations of our network to the frequency range of interests.

Once we predict the enhanced phasor field at the aperture, the hidden scenes can be reconstructed using the existing wave propagation operators.
We utilize the 2D fast Fourier transform (FFT) based RSD algorithm \cite{liu2020diffraction} in this work.

%% file: Main/Partials/figure_method_frequency.tex
\begin{figure}[t!]
    \centering
    \includegraphics[trim={0 20 0 0pt}, width=\linewidth]{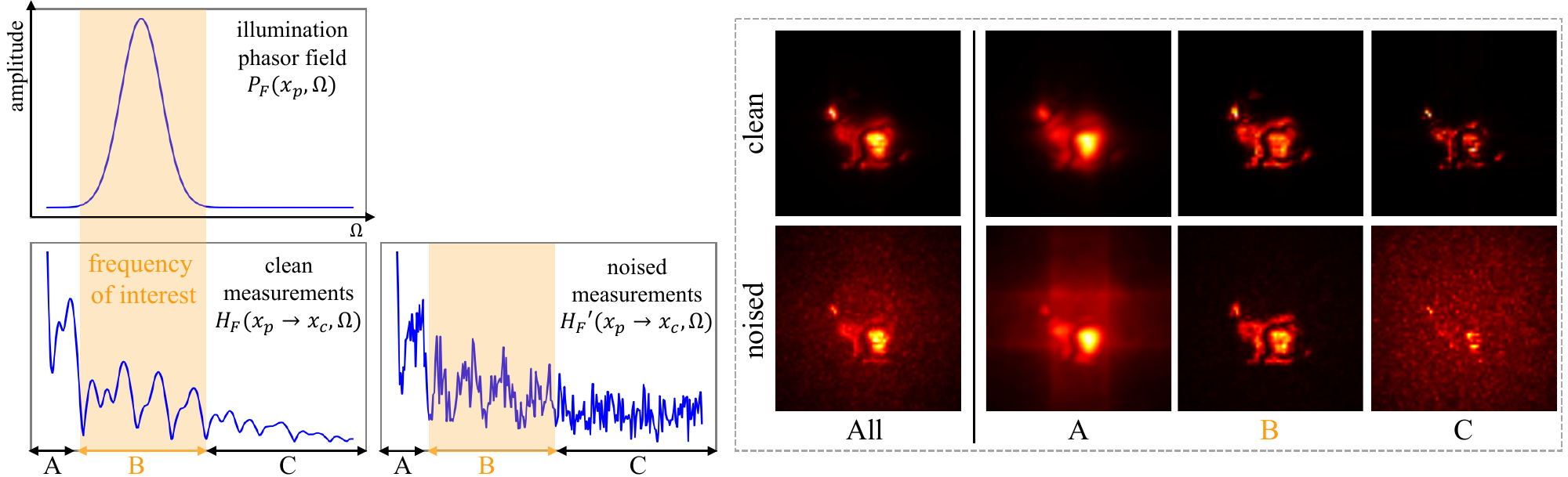}
    \caption{\textbf{(left)} The illumination function in the frequency domain (top), and amplitudes of measurements of the Stanford bunny, both clean and with Poisson noise (bottom). Signals at the center pixel are visualized. \textbf{(right)} Reconstruction results of FK \cite{lindell2019fk} on frequency-filtered measurements. Informative signals are mostly observed in a certain frequency range, whereas the Poisson noise affects across the entire spectrum \cite{hernandez2017spad}.}
    \label{fig:method_frequency}
\end{figure}

%% file: Main/Partials/figure_method.tex
\begin{figure*}[t]
    \centering
    \includegraphics[trim={0 15 0 0pt}, width=1\linewidth]{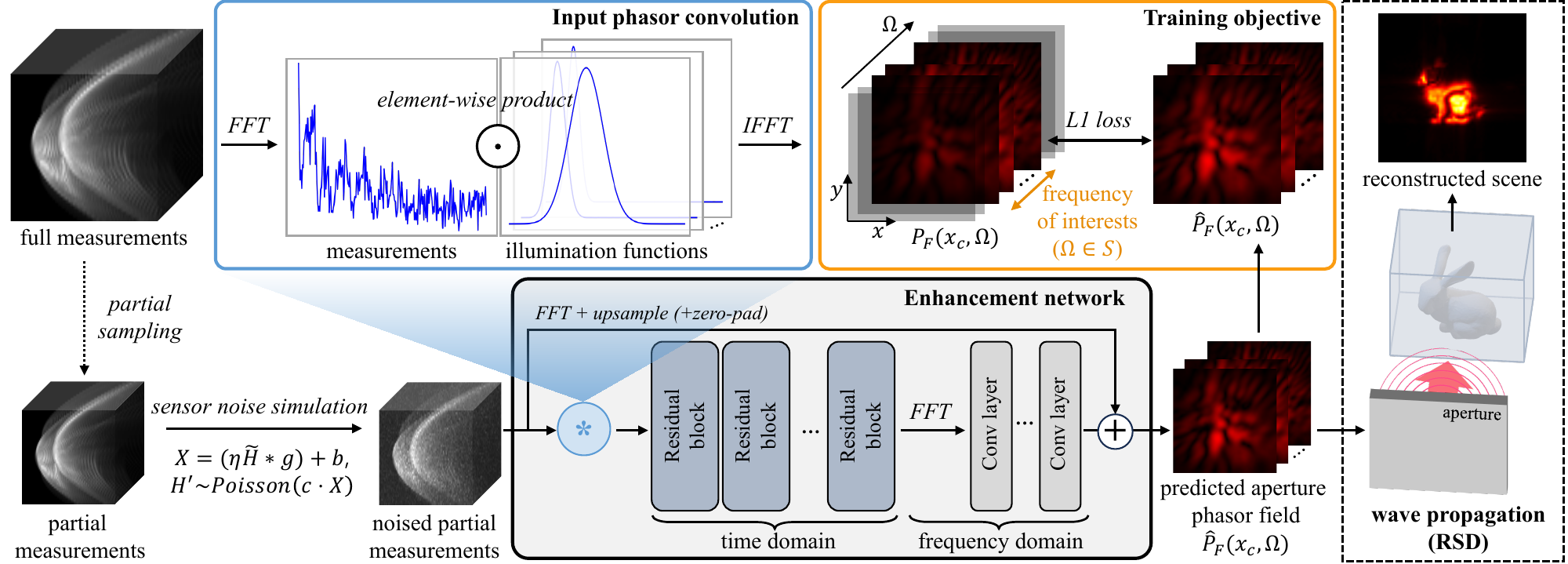}
    \caption{
    The overview of the proposed LEAP. Our model takes noisy partial measurements and learns to predict clean and complete phasor wavefronts at the aperture. Hidden scenes are reconstructed by propagating the predicted phasor field with RSD \cite{liu2020diffraction}.
    }
   \label{fig:method}
\end{figure*}

%% file: Main/Sections/experiment.tex
\section{Experiment}
\label{sec:experiment}
To demonstrate the effectiveness of the proposed method, we conduct the experiments in two practical acquisition scenarios: (1) sparse sampling and (2) scanning with a smaller aperture, of which scanning patterns are are depicted in \Fref{fig:intro} (b). 
Details of the experimental setup can be found in Supplement C, D.

\textblock{Evaluation scenarios.}
The acquisition scenarios are more specifically divided into 4 setups: confocal sparse scanning with 16 $\times$ 16 samplings (denoted as Conf-16), confocal sparse scanning with 8 $\times$ 8 points (Conf-8), confocally scanning the smaller area with size \sqsize{1}{m}{1}{m} and $16 \times 16$ samplings (Conf-small), and non-confocal sparse scanning of $16 \times 16$ points (Non-16).

Following previous works \cite{ye2021compressed, wang2023ssn, liu2023sscr, li2024usm}, we assess the performance in partial sampling scenarios with the subset of the measurements, of which scanning points are evenly sampled from full measurements with appropriate spatial strides (and center-crop for the smaller aperture case). 
We aim to recover target measurements with \sqsize{2}{m}{2}{m} apertures, 32 ps bin resolution and $64 \times 64$ samplings, which are then used to reconstruct $64 \times 64 \times 64$ hidden volumes.

\textblock{Baselines.}
We compare our method with several representative baselines:
FK \cite{lindell2019fk}, LCT \cite{o2018lct}, RSD \cite{liu2020diffraction} with nearest and trilinear interpolation methods, SSCR \cite{liu2023sscr} as an optimization-based few-shot NLOS method, and LFE \cite{chen2020lfe}, USM \cite{li2024usm}, SSN \cite{wang2023ssn} as learning-based baselines.
SSN and USM are designed to solve NLOS imaging with partial measurements: SSN recovers missing signals from partial measurements, while USM extends the architecture of LFE, consisting of the signal recovery network, the feature propagator, and the volume refinement module.

We follow the original paper to reproduce SSN \cite{wang2023ssn} as their codes are not available.
Results of the learning-based methods are reproduced with our synthetic dataset.
For a fair comparison, LFE and USM are modified to employ RSD as a propagator, are trained using our noise augmentation and supervised with 2D labels generated by projecting the outputs of RSD using the optimal measurements.
We refer to Supplement E for more baselines, details, and results.

\textblock{Implementation detail.}
Our model is implemented with PyTorch and trained 160 epochs using a single RTX A5000 GPU, which takes less than a day in our environment.
We employ 7 wavelengths for the input phasor field convolution, $\gamma = 0.1$, and the target wavelength $\lambda_T = 9.375$ cm.
We deliver 2D projected results of all methods, our model and baselines, obtained with maximum intensity projection. More implementation details are described in Supplement D.

\input{Main/Partials/table_quantitative}
\subsection{Synthetic Dataset Evaluation}
To train and validate our model, we generate a synthetic NLOS dataset from ShapeNet \cite{shapenet2015} using the NLOS renderer of \cite{chen2020lfe}.
We use 15,000 objects from all categories for generation, 11,000 objects for training and 4,000 objects for validation.
The generated synthetic dataset consists of measurements with \sqsize{2}{m}{2}{m} scan area, 64 $\times$ 64 sampling points and 32 ps bin resolution with time jitter.

For quantitative comparisons, We measure peak-signal-to-noise ratio (PSNR) and structural similarity index (SSIM) for the visual quality, and root-mean-square error (RMSE) for the accuracy of the reconstructed geometry.
The 2D projected results of RSD with the optimal measurements are served as the ground truth intensity images.
Due to the variance of reconstructed albedo values, we compare with methods based on RSD: RSD with nearest (RSD\textsubscript{Nearest}) and trilinear interpolations (RSD\textsubscript{Linear}), and learning-based methods.
Measurements with the sensor noise model in \Sref{sec:leap} are used for the evaluation.

\textblock{Results.}
\Tref{table:quantitative} reports the quantitative results on the synthetic dataset in all 4 evaluation scenarios.
The proposed model outperforms all other methods in both terms of visual quality and geometry.
RSD with both interpolation methods produce worst results.
SSN \cite{wang2023ssn} fails to learn robust representations from noisy measurements and yields inaccurate results.
LFE also fails to deliver meaningful results, indicating difficulties in exploiting learned priors from inaccurately propagated feature volumes (see results of RSD).
By exploiting the signal recovery network on top of LFE, USM achieves improvement compared to LFE and delivers the second best results.
Nevertheless, the performance gap between USM and our model indicates that our phasor-based scheme guides our model to effectively extract informative signals from noisy partial inputs.
For more results on the synthetic dataset, please refer to Supplement F.1.


\input{Main/Partials/figure_confocal}
\subsection{Confocal Real-World Evaluation}
To evaluate the generalization capability to real-world measurements, we compare the results on the Stanford confocal real-world dataset \cite{lindell2019fk}.
We choose Bike and Dragon instances for the evaluation, which have lower photon counts and SNR compared to other instances, making them more challenging to reconstruct.
The original measurements of Stanford dataset are captured with $512 \times 512$ samplings, \sqsize{2}{m}{2}{m} apertures and 32 ps bin resolution.
We first downsample the measurements by $2\times$, which results in slight increase of the exposure time per pixel.
Then we sub-sample the partial measurements with spatial strides and center crop.
We use measurements with approximately 55 ms exposure per pixel, which corresponds to the 60 minutes exposure time of the original ones.
Results of the trilinear interpolation and LFE are omitted in this section due to the space constraints, and they can be found in Supplement E.3.

\input{Main/Partials/figure_nonconfocal}
\textblock{Sparse sampling.}
Results in the $16 \times 16$ sparse sampling scenario are reported in \Fref{fig:qualitative_confocal} (first, fourth rows).
Our model outperforms all other baselines, producing clean results with details.
The inverse NLOS methods with the nearest interpolation only reconstruct coarse shapes of the objects, most of which are hardly identifiable due to the artifacts.
SSCR only reconstructs some parts of the objects.
SSN fails to learn noise-robust representations due to the absence of denoising, leading to outputs with severe artifacts.
USM produces plausible outputs in general, yet some artifacts can be observed in its results.
In contrast, our model successfully reconstructs clean shapes of the objects with fine details, \eg, the rear wheel of Bike and the head of Dragon.

The efficacy of our method is more highlighted with $8 \times 8$ sparse sampling, having $4 \times$ shorter scanning time than $16 \times 16$.
As shown in the second, fifth rows of \Fref{fig:qualitative_confocal}, none of the baselines deliver plausible results.
On the other hand, our method delivers the results where several parts of the objects, \eg the wheels of Bike and the legs of Dragon, are clearly visible.
Our results in sparse sampling scenarios showcase the effectiveness of our phasor-based network under real-world noise, leading to significant improvement compared to other baselines.


\textblock{Smaller aperture.}
Our model also achieves high-quality results in the smaller aperture case, as shown in \Fref{fig:qualitative_confocal} (third, sixth rows).
Results of the nearest interpolation only contain coarse shapes of the objects near the aperture with severe artifacts.
Again, SSCR misses many parts of the objects, USM produces distorted shapes of the objects, and SSN suffer from the effects of the noise.
The results of our method, in which many parts of the objects placed out of the aperture are reconstructed (the wheels of Bike), demonstrate that the proposed model can realize applications with limited apertures.

\subsection{Non-Confocal Real-World Evaluation}
Finally, we evaluate our method in the non-confocal $16\times 16$ sparse sampling scenario.
We compare the results on the measurements of Resolution provided in \cite{liu2020diffraction}, captured with \sqsize{1.8}{m}{1.3}{m} apertures, 1 cm sampling distance, 4 ps bin resolution and 1 s exposure per pixel.
We first apply spatial zero-pad and temporal averaging to the measurements to make \sqsize{2}{m}{2}{m} apertures and 32 ps bin resolution.
Then the partial inputs are sub-sampled with spatial strides.

As in \Fref{fig:qualitative_nonconfocal}, our model exhibits much improved performance, showing the results closes to the results of RSD with full measurements.
On the other hand, results of RSD with the interpolation methods are blurry and contain some artifacts.
LFE only reconstructs coarse shapes of the object.
Interestingly, all learning-based methods employing signal recovery networks (USM, SSN, Ours) seem to produce compelling results under the low-noise condition, where the exposure time per pixel is sufficient (about $20\times$ longer than the Stanford dataset) and the target objects with high-reflectivity.
We further probe effects of the denoising criterion and noise robustness of the models in the following sections, and provide more results in the sparse non-confocal scenario in Supplement F.6.

\input{Main/Partials/figure_ablation_denoising}
\input{Main/Partials/figure_ablation_frequency}
\subsection{Ablation Study}
\label{sec:ablation}
To verify the effectiveness of the proposed concepts, we conduct the ablations in both confocal and non-confocal $16 \times 16$ sparse sampling scenarios.

\input{Main/Partials/table_ablation}
\textblock{Denoising and the phasor-based network.}
We first conduct the ablation on the denoising criterion for the signal recovery problem. We compare our model with several variants of signal recovery networks: SSN \cite{wang2023ssn} and the same network trained with the denoising criterion (SSN+), our enhancement network in the time domain (w/o phasor-based scheme, Ours\textsubscript{time}) with and without denoising.
We report the results on Bike and Resolution for the real-world evaluations.

The ablation results on the synthetic dataset and the real-world measurements are reported in \Tref{table:ablation_denoising} and \Fref{fig:ablation_denoising}.
Adding denoising criterion greatly helps the network to remove background noise and learn more robust representations.
However, as shown in \Fref{fig:ablation_denoising}, applying denoising criterion often results in over-smoothed outputs (Resolution) and missing details of the objects (the rear wheel of Bike).
By applying the phasor-based scheme, our model achieves both quantitative and qualitative improvements, producing clean results with fine details of the objects.
Notably, attention branches employed in SSN make no meaningful improvement but rather results in degraded outputs and expensive computations compared to our model.
Based on these results, we conclude that the \naive application of the denoising often leads to the unfavorable results, highlighting the effectiveness of our phasor-based network.

\textblock{Frequency filtering.}
Next, we explore the effects of frequency filtering by modifying the target illumination function.
We compare with two illumination functions: (1) passing all frequencies smaller than the central frequency $\Omega_C$ (low-pass), (2) passing all frequencies higher than $\Omega_C$ (high-pass).
Shapes of each illumination function are shown in \Fref{fig:ablation_frequency}, with their frequency ranges closely related to those in \Fref{fig:method_frequency}.
These models also employ additional wavelengths in the input phasor convolution to sufficiently cover the target frequency ranges.

As reported in \Tref{table:ablation_frequency} and \Fref{fig:ablation_frequency}, both low-pass and high-pass models deliver degraded results compared to our model.
Interestingly, the low-pass model delivers worse results than the high-pass model, failing to reconstruct details of the objects in the real-world (see \Fref{fig:ablation_frequency}, the head of Dragon).
This also corresponds to the observations made in prior works \cite{rahaman2019spectral, fridovich2022spectral}, which discovered the spectral bias of neural networks towards lower-frequency signals.
Such results demonstrate that employing phasor wavefronts as band-limited signals for the supervision guides our network to effectively avoid the spectral bias.

%% file: Main/Partials/table_quantitative.tex
\begin{table}[t!]
\setlength{\tabcolsep}{2pt}
\centering
\small
\caption{
Quantitative results on the synthetic dataset.
RMSE values of USM \cite{li2024usm} are omitted as its original version does not include depth map reconstruction.
}
\resizebox{\columnwidth}{!}{
\begin{tabular}{c|ccc|ccc|ccc|ccc}
\toprule
\multirow{2}{*}{Method} & \multicolumn{3}{c}{Conf-16}& \multicolumn{3}{c}{Conf-8} & \multicolumn{3}{c}{Conf-small}& \multicolumn{3}{c}{Non-16} \\
 & PSNR$\uparrow$ & SSIM$\uparrow$ & RMSE$\downarrow$ & PSNR$\uparrow$ & SSIM$\uparrow$ & RMSE$\downarrow$ & PSNR$\uparrow$ & SSIM$\uparrow$ & RMSE$\downarrow$ & PSNR$\uparrow$ & SSIM$\uparrow$ & RMSE$\downarrow$ \\
\midrule
RSD\textsubscript{Nearest} & 14.85 & 0.1515 & 0.8232 & 12.65 & 0.0855 & 0.8919 & 19.73 & 0.3743 & 0.3073 & 19.67 & 0.3218 & 0.5020 \\
RSD\textsubscript{Linear} & 14.62 & 0.1631 & 0.7536 & 11.28 & 0.0760 & 0.8884 & 19.90 & 0.4664 & 0.2046 & 19.29 & 0.3224 & 0.4856 \\
LFE \cite{chen2020lfe} & 23.05 & 0.6729 & 0.2247 & 17.45 & 0.4098 & 0.3282 &22.92 & 0.6826 & 0.2852 & 29.47 & 0.8077 & 0.2898 \\
USM \cite{li2024usm} & 29.99 & \textbf{0.8994} & - & 25.75 & 0.8235 & - & 26.94 & 0.8519 & - & 35.14 & 0.9313 & - \\
SSN \cite{wang2023ssn} & 23.27 & 0.4506 & 0.2699 & 21.49 & 0.4426 & 0.2259 & 21.98 & 0.4629 & 0.2036 & 29.45 & 0.6367 & 0.1798 \\
Ours & \textbf{32.02} & 0.8949 & \textbf{0.0892} & \textbf{28.07} & \textbf{0.8472} & \textbf{0.0962} & \textbf{28.31} & \textbf{0.8556} & \textbf{0.0969} & \textbf{37.45} & \textbf{0.9625} & \textbf{0.1414} \\
\bottomrule

\end{tabular}

}
\label{table:quantitative}
\end{table}

%% file: Main/Partials/figure_confocal.tex
\begin{figure*}[t]
    \centering
    \includegraphics[trim={0 20 0 0pt}, width=1\linewidth]{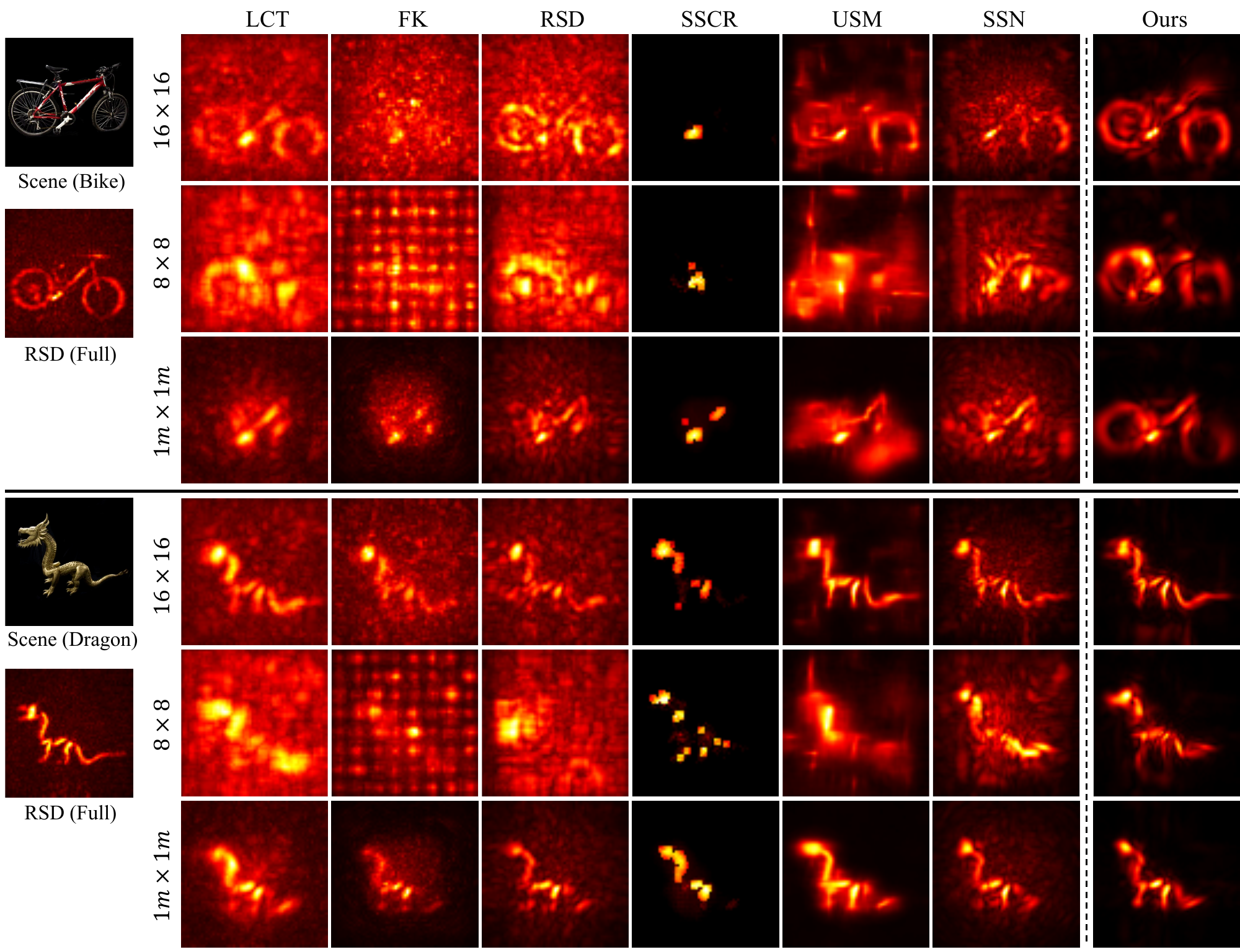}
    \caption{
    Qualitative results on Bike, Dragon from the Stanford real-world dataset \cite{lindell2019fk}.
    We report the results of FK, LCT, and RSD with the nearest interpolation. Evaluation scenarios involve $16 \times 16$ and $8 \times 8$ sparse samplings with \sqsize{2}{m}{2}{m} apertures, and the \sqsize{1}{m}{1}{m} smaller aperture with $16 \times 16$ samplings.
    }
    \label{fig:qualitative_confocal}
\end{figure*}

%% file: Main/Partials/figure_nonconfocal.tex
\begin{figure}[t]
\centering
    \includegraphics[trim={0 20 0 0pt}, width=1\linewidth]{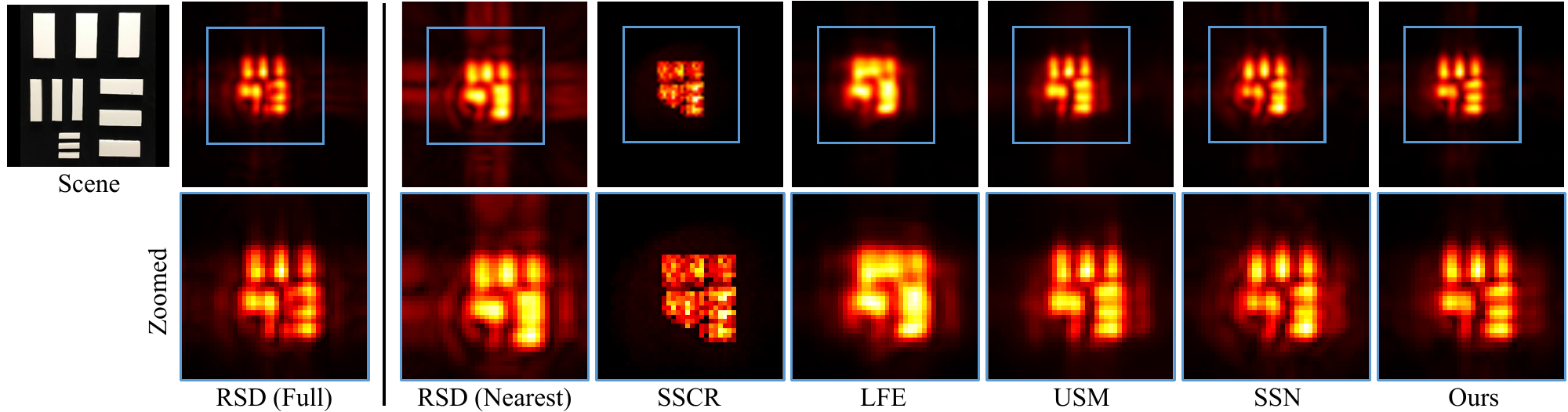}
    \caption{
    Qualitative results of non-confocal $16 \times 16$ sparse samplings, on Resolution of the real-world dataset \cite{liu2020diffraction}. All methods employing signal recovery networks produce plausible results with sufficiently long exposure time and white diffuse objects.
    }
   \label{fig:qualitative_nonconfocal}
\end{figure}

%% file: Main/Partials/figure_ablation_denoising.tex
\begin{figure}[t]
    \centering
    \includegraphics[trim={0 20 0 0pt}, width=1\linewidth]{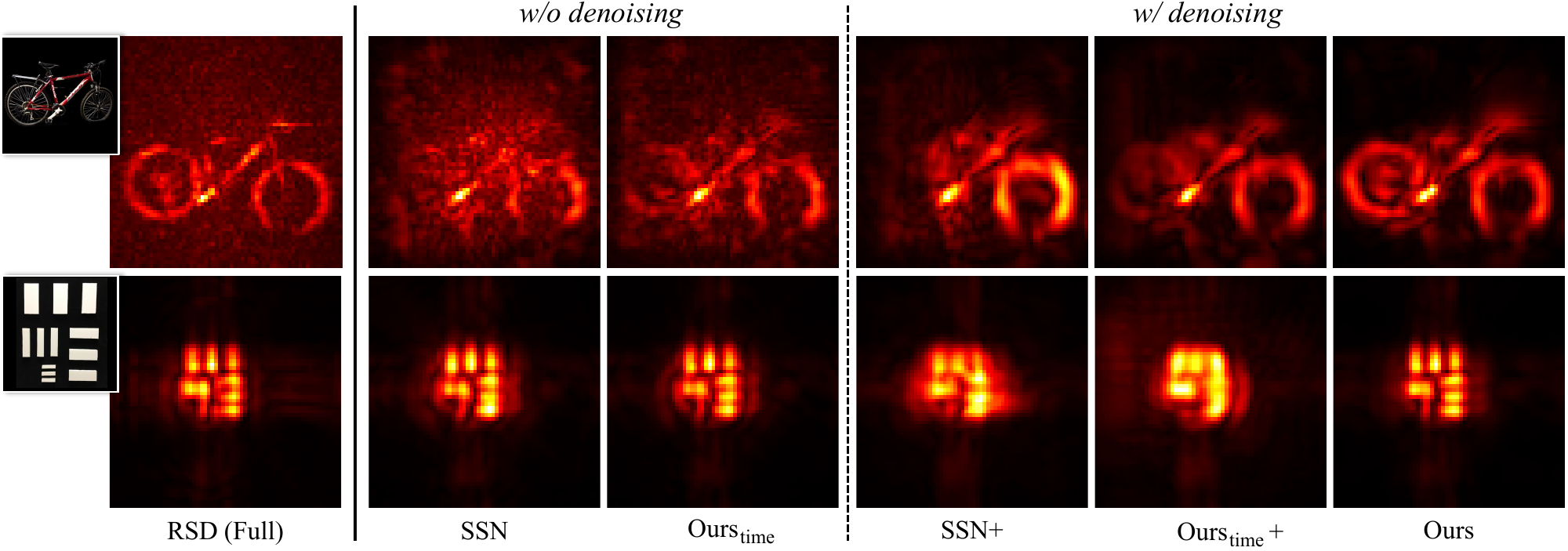}
    \caption{
    Qualitative ablation results on the denoising criterion. Results in $16 \times 16$ sparse sampling scenarios under both confocal (top) and non-confocal (bottom) setups are reported. While the \naive addition of denoising effectively reduces background noise, it often yields over-smoothed and degraded results with missing details.
    }
   \label{fig:ablation_denoising}
\end{figure}

%% file: Main/Partials/figure_ablation_frequency.tex
\begin{figure}[t]
    \centering
    \includegraphics[trim={0 20 0 0pt}, width=1\linewidth]{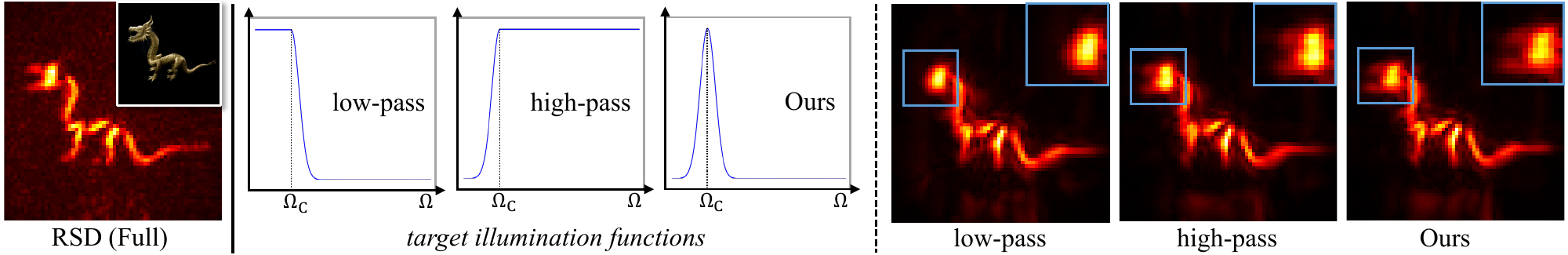}
    \caption{
    Qualitative ablation results on the frequency filtering. Left plots indicate the modified target illumination function used for the models. Results on $16 \times 16$ sparsely sampled confocal measurements of Dragon are reported.
    }
   \label{fig:ablation_frequency}
\end{figure}

%% file: Main/Partials/table_ablation.tex
\begin{table}[t]
\begin{adjustbox}{valign=t}

\begin{minipage}{0.49\linewidth}
\setlength{\tabcolsep}{3pt}
\centering
\caption{
Ablation results on the denoising criterion. `+' indicates that the models (SSN+, Ours\textsubscript{time}+) are trained with the denoising criterion.
}
\label{table:ablation_denoising}
\resizebox{\columnwidth}{!}{
\begin{tabular}{c|cc|cc}
\toprule
\multirow{2}{*}{Method} & \multicolumn{2}{c}{Conf-16}& \multicolumn{2}{c}{Non-16} \\
 & PSNR$\uparrow$ & RMSE$\downarrow$ &  PSNR$\uparrow$ & RMSE$\downarrow$ \\
\midrule
SSN \cite{wang2023ssn}  & 23.27 & 0.2699 & 29.45 & 0.1798 \\
Ours\textsubscript{time} & 23.85 & 0.2352 & 29.89 & 0.1736 \\
SSN+ & 29.55 & 0.0949 & 35.47 & 0.1435 \\
Ours\textsubscript{time}+ & 30.69 & 0.0924 & 36.36 & 0.1423 \\
Ours & \textbf{32.02} & \textbf{0.0892} & \textbf{37.45} & \textbf{0.1414} \\
\bottomrule
\end{tabular}
}
\end{minipage}
\end{adjustbox}
\hfill
\begin{adjustbox}{valign=t}
\begin{minipage}{0.49\linewidth}
\setlength{\tabcolsep}{3pt}
\centering
\caption{
Ablation results on the phasor-based frequency filtering. The ``low-pass'' model passes frequencies lower than the central frequency $\Omega_C$, and the ``high-pass'' model passes frequencies higher than $\Omega_C$.
}
\label{table:ablation_frequency}
\resizebox{\columnwidth}{!}{
\begin{tabular}{c|cc|cc}
\toprule
\multirow{2}{*}{Method} & \multicolumn{2}{c}{Conf-16}& \multicolumn{2}{c}{Non-16} \\
 & PSNR$\uparrow$ & RMSE$\downarrow$ &  PSNR$\uparrow$ & RMSE$\downarrow$ \\
\midrule
low-pass & 31.23 & 0.0903 & 36.72 & 0.1420 \\
high-pass & 31.54 & 0.0909 & 37.09 & 0.1422 \\
Ours & \textbf{32.02} & \textbf{0.0892} & \textbf{37.45} & \textbf{0.1414} \\
\bottomrule
\end{tabular}
}
\end{minipage}
\end{adjustbox}
\end{table}

%% file: Main/Sections/discussion.tex
\section{Analysis and discussion}
\label{sec:discussion}
\textblock{Runtime analysis.}
Our network takes $20.7\ \textrm{ms}$ for processing measurements with $16 \times 16$ samplings, and the GPU version of RSD takes 3.9 ms to reconstruct $64 \times 64 \times 64$ volumes.
On the other hand, SSN employs computationally expensive attention branches and thus takes 72.9 ms to process $16 \times 16$ inputs.
As a result, our entire pipeline takes less than 25 ms, highlighting the suitability of our method for real-time applications.
The latency is measured using a single RTX 3090 GPU.
Please refer to Supplement E.3 for comparisons with more baselines.

\input{Main/Partials/figure_exposure}
\textblock{Noise robustness of learning-based methods.}
To examine the noise robustness of learning-based methods, we report the results on Bike with shorter and longer exposure times per pixel.
As shown in \Fref{fig:exposure}, all methods produce plausible results with sufficiently long exposure time.
While the denoising criterion seems less effective in such low-noise conditions, its effects are clearly highlighted when the exposure time (and thereby the total scanning time) is reduced.
Simple addition of denoising to SSN (SSN+) or incorporating with the volume refinement network (USM) fail to reveal some details of the object, \eg the wheels of Bike.
Our model consistently deliver high-quality results and presents its noise robustness with a shorter exposure time and a shorter scanning time.

\textblock{Incorporating with other NLOS methods.}
While our model predicts the optimal phasor wavefronts at the aperture in the frequency domain, we can still incorporate with other inverse NLOS methods with slight modification.
We refer to Supplement F.3 for more information and results with other NLOS methods.

\textblock{Limitation and future works.}
Since the proposed LEAP exploits learned priors from the synthetic dataset, the performance of our model is affected by the quality of the generated samples.
This can be further improved by using more complicated objects and precise NLOS renderer \cite{galindo19znlos}.
In addition, incorporating recent advances in generative diffusion models \cite{ho2020ddpm, song2019generative, song2020score} would also be an interesting direction for the future works.


%% file: Main/Partials/figure_exposure.tex
\begin{figure}[t]
    \centering
    \includegraphics[trim={0 20 0 0pt}, width=1\linewidth]{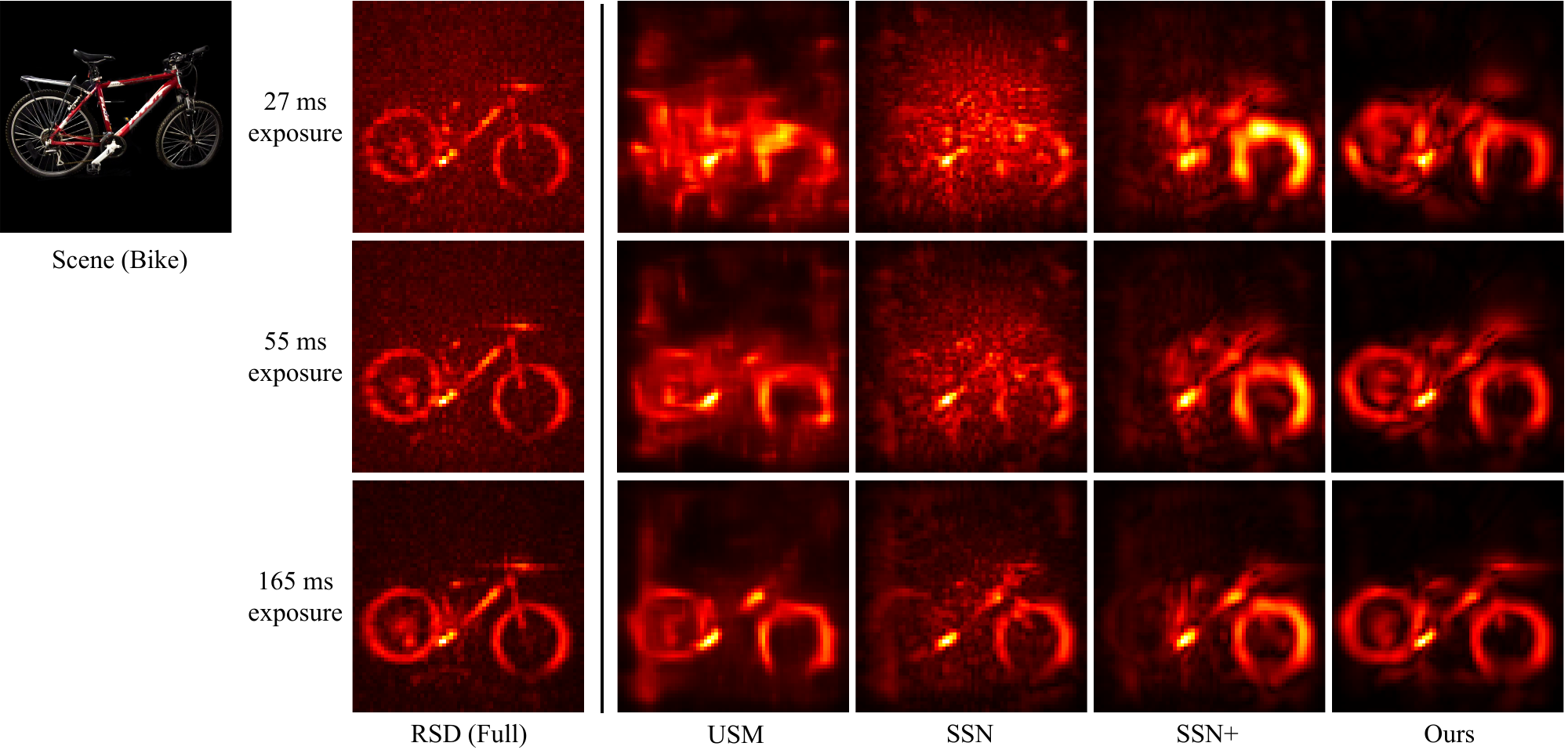}
    \caption{
    Results on sparse $16 \times 16$ confocal measurements of Bike \cite{lindell2019fk}, with shorter ($27\ \textrm{ms}$ per pixel) and longer ($165\ \textrm{ms}$ per pixel) exposure time.
    The denoising criterion becomes evidently effective as the exposure time is reduced, but it fails to reveal some details of the object. Our method consistently presents high-quality results, exhibiting the noise robustness of the proposed phasor-based scheme.
    }
   \label{fig:exposure}
\end{figure}

%% file: Main/Sections/conclusion.tex
\section{Conclusion}
We proposed the learning-based method LEAP, which learns to enhance noisy partial measurements and enables NLOS imaging with fewer samplings and smaller apertures.
Our enhancement network with the phasor-based scheme presented its effectiveness in various scanning scenarios, showing high-fidelity results while being robust to the noise.
We believe our method can serve as an effective solution to address the exhaustive scanning procedures in NLOS imaging.

\section*{Acknowledgements} 
This work was supported by the Samsung Research Funding Center (SRFC-IT2001-04), Artificial Intelligence Innovation Hub under Grant RS-2021-II212068, and an Institute of Information \& communications Technology Planning \& Evaluation (IITP) grant funded by the Korean Government (MSIT) (No. RS-2020-II201361, Artificial Intelligence Graduate School Program (Yonsei University)).

%% file: Supple/supple.tex
\appendix

In this supplementary material, we provide additional details related to the topics discussed in the manuscript.
The contents of the supplementary material are as follows:
\begin{itemize}[leftmargin=0.7cm]
    \item Details of the analysis of frequency components and effects of noise, discussed in Section 3 of the manuscript (\Sref{sec:frequency_analysis}).
    \item Details of our method including network architecture, noise parameters, and wave propagation for the reconstruction (\Sref{sec:method_detail}).
    \item Details of the dataset (\Sref{sec:dataset_detail}).
    \item Details of the experiment setup (\Sref{sec:experiment_setup_detail}).
    \item Details of baselines and additional comparison with more baseline methods (\Sref{sec:baseline_detail}).
    \item Additional experimental results (\Sref{sec:additional_result}).
    \item Further discussion on societal impact, supervision of the reconstructed volumes, imaging systems, and neural networks for NLOS imaging (\Sref{sec:additional_sensors}).
    \item Error bar to validate the reproducibility of the proposed method (\Sref{sec:error_bar}).
\end{itemize}

\input{Supple/Sections/frequency_analysis}
\input{Supple/Sections/method_detail}
\input{Supple/Sections/dataset_detail}
\input{Supple/Sections/experiment_detail}
\input{Supple/Sections/baseline_detail}
\input{Supple/Sections/additional_result}
\input{Supple/Sections/additional_discussion}
\input{Supple/Sections/error_bar}

%% file: Supple/Sections/frequency_analysis.tex
\section{Frequency Components Analysis Detail}
\input{Supple/Partials/figure_frequency}
\label{sec:frequency_analysis}
In Section 3.2 of our main paper, we discovered the effects of noise across the frequency components by visualizing reconstruction results of FK \cite{lindell2019fk}, one of the representative NLOS methods.
These reconstruction procedures include a band-pass filter that discards signals outside a specific frequency range (\ie, for a range B, a band-pass filter retains all frequency components within the range B and discard all other components).
For better comprehension, to visualize signals in Figure 1, we omitted some of the lowest frequency components having too large amplitudes (10 lowest frequency components).

Here, we further visualize reconstruction results on frequency-filtered measurements using various NLOS methods: back-projection (BP), LCT \cite{o2018lct}, and RSD \cite{liu2020diffraction}.
We also provide results of these methods on real-world measurements of Statue \cite{lindell2019fk}, to better discover the effects of noise and frequency filtering.
As shown in \Fref{fig:supple_frequency_analysis}, in most cases, all methods produce the noise-robust results with details of the objects in when incorporated with the band-pass filter retaining the range B (near the central frequency).

%% file: Supple/Partials/figure_frequency.tex
\begin{figure*}[t]
    \begin{center}
        \includegraphics[width=1\linewidth]{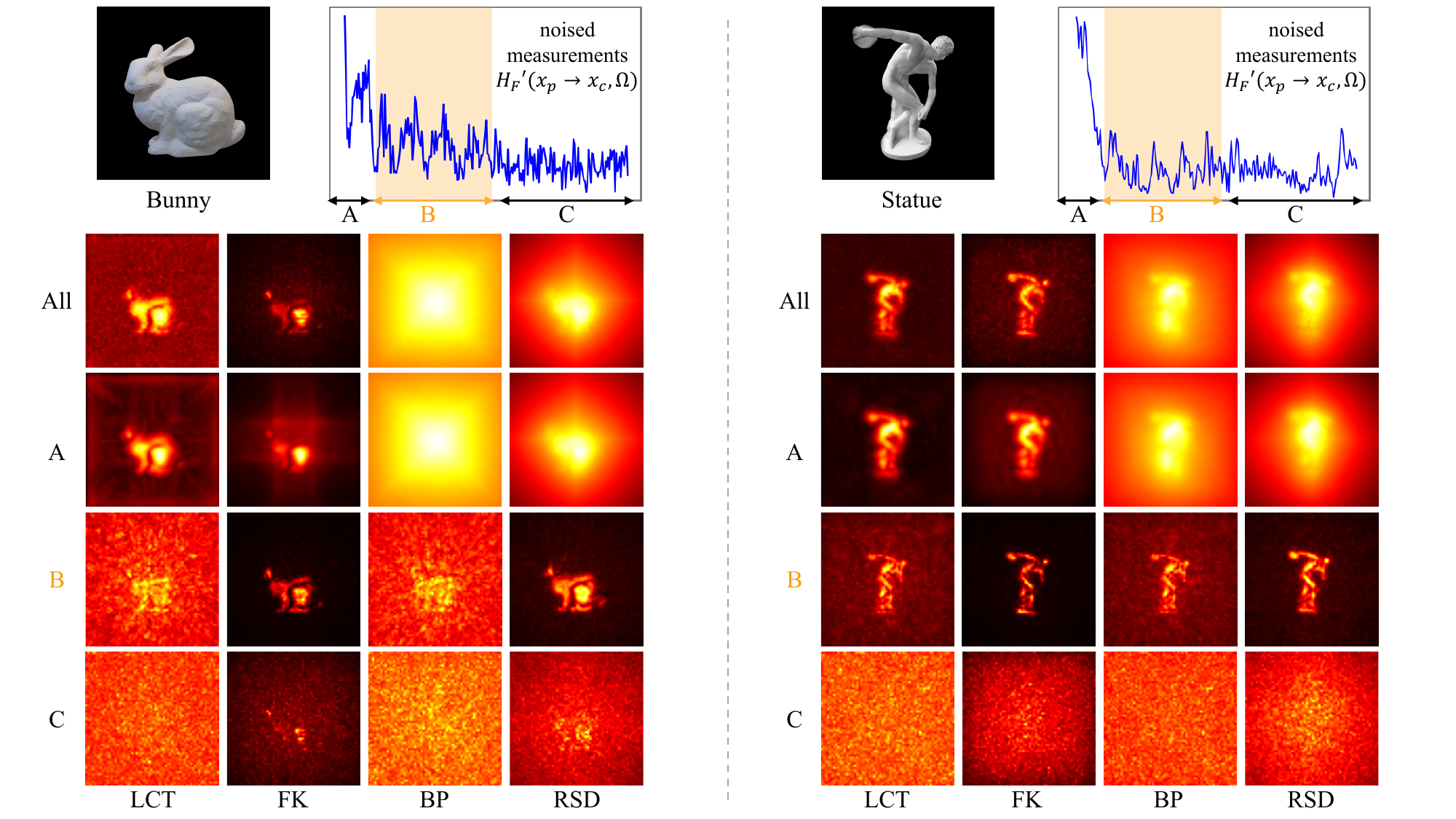}
    \end{center}
    \vspace{-10pt}
    \caption{
    Extended visualization results on frequency-filtered measurements. \textbf{(left)} Reconstruction results on noised measurements of Stanford bunny using LCT \cite{o2018lct}, FK \cite{lindell2019fk}, BP, and RSD \cite{liu2020diffraction}. \textbf{(right)} Reconstruction results on real-world measurements of Statue \cite{lindell2019fk} using LCT \cite{o2018lct}, FK \cite{lindell2019fk}, BP, and RSD \cite{liu2020diffraction}.
    Measurements within the frequency range B (around the central frequency) yield noise-robust reconstruction results with fine details of the objects in most cases.
    }
    \label{fig:supple_frequency_analysis}
\end{figure*}

%% file: Supple/Sections/method_detail.tex
\section{Method Details}
\label{sec:method_detail}

\subsection{Noise Model Parameters}
As described in the main paper, we follow the computational model of SPAD used in \cite{hernandez2017spad, chopite2020deep, chen2020lfe}.
We employ a simplified version of SPAD model, similar with \cite{chen2020lfe, chopite2020deep}, which ignore the effects of cross-talk and afterpulsing.
For the photon detection efficiency $\eta$, it is defined to ensure the top 10,000 measurement histograms with the highest values to contain at most 100 average detected photons:
\begin{equation}
  \eta =\begin{cases}
    \frac{100}{m}, & \text{if $m > 100$}. \\
    1, & \text{otherwise}.
  \end{cases}
\end{equation}
Here, $m$ is the average of the top 10,000 histograms of the measurements having highest values.
We compute the background noise ratio $d$ by multiplying the maximum photon counts of measurements and random values sampled from the range $[0.1, 0.2]$. The exposure term $c$ controls the number of scans accumulated to obtain the final measurements. Smaller values of $c$ lead to the greater effects of Poisson noise, as the variance of Poisson distribution is determined by $c$. We randomly sample $c$ from the range $[0.1, 1]$.

\subsection{Network Architecture}
Given partially sampled measurements, our model first temporally crops the measurements by choosing the region with size 256 where the sum of the photons is maximum.
Then the measurements are convolved with multiple illumination wavefronts, of which wavelength coefficients are $\{0.8, 0.9, 1.0, 1.25, 1.5, 2.0, 2.5\}$.
The corresponding wavelengths are obtained by multiplying each coefficient with $2\Delta_p$, where $\Delta_p$ is the sampling distance.
We rescale the illumination wavefronts by dividing them with their maximum value, resulting in coefficients within the range $[0, 1]$.
Components with amplitudes smaller than $\gamma = 0.1$ are then discarded.
By concatenating all real and imaginary components, our enhancement network takes measurement volumes with size $256 \times N' \times N'$ and the feature dimension 14, where $N'$ is the spatial resolution of partially sampled inputs.

The enhancement network consists of ten 3D residual blocks. Each residual block has 2 3D convolution layers with a kernel size 3 and one convolution layer with a kernel size 1 as a shortcut. We concatenate 3D position information $(x,y,t)$ of each bin at the beginning of each residual block.
In seventh and ninth residual blocks, we upsample the feature volumes by $(4, 1)$ and $(2, 1)$ when $N' = 16$, and $(4, 2)$, $(2, 1)$ when $N' = 8$, where the first denotes the temporal and the second denotes the spatial upsample factor.
The upsample operations in these blocks are performed using a pixel shuffle.
Next, the FFT operator is applied to convert the feature volumes into the frequency domain. The last three convolution layers with a kernel size $(1, 3, 3)$, along with the pixel shuffle operator, are then used to predict the residuals in the frequency domain. Both the real and imaginary parts of the residuals are added to the upsampled input measurements in the frequency domain. The upsampling is achieved through nearest interpolation, with zero-padding applied for cases with a smaller aperture, resulting in a spatial resolution of $64 \times 64$.
Finally, we predict the aperture wavefront $\hat{\mathcal{P}}_{\mathcal{F}}(\mathbf{x_c}, \Omega)$ as we described in the main paper.
Note that we can obtain the time domain predictions of the measurements if the final convolution with the target illumination function $\mathcal{P}(x_p, \Omega)$ is omitted.
We deliver the results using the time domain predictions with other inverse NLOS methods in \Sref{sec:incorporating_others}.

For the models used in the ablation study (Ours\textsubscript{time}), we remove the last FFT operator to predict the optimal measurements in the temporal domain while keeping remaining components and configurations same.
The loss is minimized in the temporal domain for these models, but we apply the same FFT operator, illumination wave coefficients with $\lambda_T$ and the filtering process to their predictions during the inference to ensure fair comparisons.

\subsection{Wave Propagation}
After our model predicts the optimal aperture wavefronts, hidden volumes are reconstructed from these predictions using the convolutional 2D FFT-based RSD algorithm \cite{liu2020diffraction}.
The propagation of the phasor field at the aperture plane $\mathbf{x_c} = (x_c, y_c, 0)$ to the hidden scenes $\mathbf{x_v} = (x_v, y_v, z_v)$ can be formulated as a 2D convolution of $\mathcal{P}(\mathbf{x_c}, \Omega)$ and the diffraction kernel $G(\cdot)$:
\begin{equation}
    \mathcal{P}_{\mathcal{F}}(\mathbf{x_v}, \Omega) = \mathcal{P}(\mathbf{x_c}, \Omega) * G(x_c, y_c, z_v, \Omega),
\end{equation}
and the 2D convolution kernel $G(x_c, y_c, z_v, \Omega)$ can be written as
\begin{equation}
    G(x_c, y_c, z_v, \Omega) = \frac{e^{-i\frac{\Omega}{c}\sqrt{x_c^2+y_c^2+z_v^2}}}{\sqrt{x_c^2+y_c^2+z_v^2}},
\end{equation}
where $c$ is the light speed.
The final hidden volumes are obtained by propagating each monochromatic wave component and then integrating them with the inverse Fourier transform:
\begin{equation}
    I(\mathbf{x_v}, t) = |\mathcal{P}_{\mathcal{F}}(\mathbf{x_v}, t)|^2 = 
    \left| \int_{\Omega_C - \Delta\Omega}^{\Omega_C + \Delta\Omega}{e^{i\Omega t} \cdot 
    \mathcal{P}_{\mathcal{F}}(\mathbf{x_v}, \Omega)\ \mathrm{\frac{d\Omega}{2\pi}}} \right|^2.
\end{equation}
For the confocal experiments, the diffraction kernel with two doubling distances is used, as in the original work \cite{liu2020diffraction}.
Since our phasor-based network directly predicts the phasor wavefronts with target wavelength $\lambda_T$ in the frequency domain, we propagate the predicted $\hat{\mathcal{P}}_{\mathcal{F}}(\mathbf{x_c}, \Omega)$ to reconstruct the hidden scenes.
For other methods, SSN \cite{wang2023ssn}, Ours\textsubscript{time}, and the interpolation method, we first compute the phasor field at the aperture using the virtual illumination function with the target wavelength $\lambda_T$ and then propagate the computed phasor wavefronts.
We empirically omit the radial drop-off term in the diffraction kernel, which we observe slightly better reconstruction quality in our experiment setups.

%% file: Supple/Sections/dataset_detail.tex
\section{Dataset Details}
\label{sec:dataset_detail}
\subsection{Synthetic Dataset}
Our synthetic dataset is generated by rendering the 3D objects in ShapeNet \cite{shapenet2015}, using the NLOS renderer provided in \cite{chen2020lfe}.
We generate two datasets, confocal and non-confocal, to train and validate our model.
Both datasets are rendered using same objects, same splits, and same augmentation parameters.
We list the parameters and details of our rendering process for future works.

\parabf{Renderer implementation and configurations.}
For more efficient rendering, we extend the NLOS renderer provided in \cite{chen2020lfe} and implement the multi-processing and multi-GPU renderer. We render our synthetic dataset using 4 RTX 2080 Ti GPU, and the rendering process takes approximately 1 day. We add time jitters to the rendered transients using the algorithm in \cite{hernandez2017spad}. We also re-implement the codes for the time jitter using PyTorch so that we can fully exploit multiple GPU resources.
Finally, Poisson sensor noise is simulated during the training process as in the main paper.

We set the size of the relay wall as $2\ \textrm{m} \times 2\ \textrm{m}$ and the target objects are placed in the cube with size $2\ \textrm{m} \times 2\ \textrm{m} \times 2\ \textrm{m}$. The renderer produces the 3D transients with size $512 \times 64 \times 64$ and the bin resolution $32\ \textrm{ps}$, and their corresponding 2D depth map with size $512 \times 512$.
We found that the quality of rendered depth maps is significantly degraded when they are rendered in low resolution. Therefore, we first render the high-resolution 2D depth maps using our OpenGL based renderer and then downsample them into $64 \times 64$.
As the colors of some objects in ShapeNet are almost black, which can possibly disrupt the training, we rescale the color of the objects to be in range $[0.3, 1]$.
Some of the samples that do not contain sufficient returning photons are excluded from the dataset.

\parabf{Random augmentation.}
For the training set, we render each object 4 times with random rotation, shift and scale.
The objects are first randomly rotated with the angles randomly sampled from the fixed ranges $[-15\degree, 15\degree]$.
After the rotations, the objects are randomly scaled by the scale factor $s$, which is sampled from the range $0.8 < s < 1.2$. The objects are then aligned to the center of the relay wall with approximately $1m$ distance along the $z$ axis, and then randomly shifted with the offsets $t$, which are sampled from the range $-0.3 < t < 0.3$.
For the validation set, we slightly reduce shift range to $-0.1 < t < 0.1$ and the ranges of rotation angles to $[-5\degree, 5\degree]$. The scale factor $s$ fixed to 1, and the objects are rendered once for the validation set.

\subsection{Real-World Dataset}
For the Stanford confocal real-world dataset \cite{lindell2019fk}, we utilize a spatial averaging by a factor 2 to slightly increase the exposure time per pixel.
This results in the full measurements with spatial resolution $256 \times 256$, from which we sub-sample partial measurements using appropriate spatial strides (16 for $16 \times 16$, 32 for $8 \times 8$, and 8 for the smaller aperture measurements).
Since we used the original measurements with total 60 minutes acquisition time in our experiments, partial measurements with $16 \times 16$ samplings and $8 \times 8$ samplings require $14.1\ \textrm{s}$ and $3.5\ \textrm{s}$ scanning time, respectively.
The $64 \times 64$ measurements used as references in our experiments require $225\ \textrm{s}$ scanning time.

For the non-confocal real-world dataset provided in \cite{liu2020diffraction}, we first apply temporal averaging to the measurements by a factor of 8.
Then we apply zero-pad with sizes 70, 20 for width and height, respectively.
This procedure yields measurements aligned with our synthetic dataset, featuring a \sqsize{2}{m}{2}{m} aperture and a $32\ \textrm{ps}$ bin resolution.
Note that, except for the alignment process described above, we did not apply any preprocessing to the original measurements.
This is to ensure proper and fair evaluations in partial sampling scenarios.

\input{Supple/Partials/table_photon_counts}
\parabf{Photon counts.}
To analyze the noise levels of real-world measurements, we assess the average photon counts of both confocal \cite{lindell2019fk} and non-confocal \cite{liu2020diffraction} measurements.
While dark counts of sensors affects across the entire histograms, actual photons returning from objects only appear in a few histograms.
Therefore, we also report the average photon counts of top-k histograms with the highest photon counts, as these are more likely to contain actual photons.
These top-k photon counts can serve as an alternative representation of the noise levels in the measurements.
As shown in \Tref{table:supp_photon_counts}, the Bike and Dragon instances lower photon counts than other instances (Statue, Teaser).
This suggests that measurements of these instances have lower albedo values and lower SNR, making the reconstruction process more challenging.
Consequently, we select Bike and Dragon instances for the evaluations presented in the main paper.
Results on additional instances are provided in \Sref{sec:other_instances}.

%% file: Supple/Partials/table_photon_counts.tex
\begin{table}[]
\setlength{\tabcolsep}{4pt}
\centering
\caption{
Average photon counts of the confocal \cite{lindell2019fk} and the non-confocal \cite{liu2020diffraction} real measurements.
`All' denotes averages across all histograms, and `top 5\%', `top 2\%', `top 1\%' denote averages for the top 5\%, 2\%, 1\% histograms with the highest photon counts.
We additionally measure the averages of these top-k histograms, as these histograms mostly contain actual photons returning from objects. 
These top-k photon counts can serve as an alternative representation for the noise levels of measurements.
We measure the photon counts of $16 \times 16$ confocal measurements with a 60 minute exposure time, of which photon counts significantly exceed the photon counts of non-confocal measurements with a 1 second exposure time.
}
\resizebox{\columnwidth}{!}{

\begin{tabular}{ccccc|cccc}
\toprule
Photon counts & \multicolumn{4}{c}{Confocal \cite{lindell2019fk}} & \multicolumn{4}{c}{Non-confocal \cite{liu2020diffraction}} \\
(avg.) & Bike & Dragon & Statue & Teaser & Resolution & ``4'' & ``NLOS'' & Shelf \\
\midrule
all & 6.1 & 5.9 & 8.46 & 11.26 & 60.5 & 131.6 & 55.1 & 49.3 \\
top 5\% & 14.8 & 15.3 & 38.3 & 47.2 & 561.2 & 872.4 & 515.9 & 327.5 \\
top 2\% & 17.4 & 18.6 & 52.0 & 65.7 & 898.0 & 1327.8 & 837.1 & 460.1 \\
top 1\% & 19.4 & 21.3 & 63.0 & 82.2 & 1179.9 & 1707.3 & 1116.1 & 565.1 \\
\bottomrule
\end{tabular}

}
\label{table:supp_photon_counts}
\end{table}

%% file: Supple/Sections/experiment_detail.tex
\section{Experiment Setup Details}
\label{sec:experiment_setup_detail}
\parabf{Labels of the synthetic dataset.}
In our pipeline, hidden volumes are reconstructed by applying the RSD algorithm \cite{liu2020diffraction} to the predictions.
Since the reconstructed albedo values vary across the inverse NLOS methods based on their formulations, namely, reconstructed intensity of hidden scenes are different between LCT \cite{o2018lct}, FK \cite{lindell2019fk}, and RSD \cite{liu2020diffraction}.
For this reason, most of the NLOS methods focus on assessing errors of the reconstructed depth maps.
Using front images as 2D labels, generated by the synthetic renderer \cite{chen2020lfe}, often fails to ensure fair comparisons, as it may introduce an advantageous influence on learning-based methods that are directly trained to predict these 2D images.

We can alternatively measure the visual quality of the reconstructed scenes by using the reconstructed intensity images, which are computed by applying the RSD algorithm to the optimal (clean and full) measurements and projecting them with a maximum intensity projection.
In this context, our comparisons involve methods based on the RSD algorithm in the manuscript. To ensure a fair assessment of visual quality without compromising the final outcomes of the methods, we slightly modify the configuration of LFE. It is adjusted to employ RSD as a feature propagator and is trained to predict the 2D projected results of RSD.
We also provide results of LFE trained with generated front images in \Fref{fig:supp_sanity}, confirming that using 2D results of RSD as a ground truth does not damage the final reconstruction quality.

\parabf{Additional implementation detail.}
To train our model, we use the AdamW \cite{loshchilov2018adamw} optimizer with weight decay 0.01, batch size 8, and learning rate 0.0001 decayed by $\times 0.1$ after 100 epochs.
For all experiments, we assume that the virtual light source starts illuminating the scenes at $t_0 = 0$.
The histograms of the measurements are aligned to start at $t_0$, when the laser first hits the relay wall.
The target modulation wavelength $\lambda_T = 9.375\ \textrm{cm}$ corresponds to the $3\Delta_p$, where $\Delta_p = 3.125\ \textrm{cm}$ is the sampling distance of the $64\times 64$ full measurements with \sqsize{2}{m}{2}{m} scan areas.
The virtual illumination function with $\lambda_T$ contains 47 frequency components in our experiment setup.
We expect positive results by using the ground truth optimal measurements with higher spatial resolution, as the illumination phasor field with shorter wavelength involves more and higher frequency components.
The input measurements are normalized by dividing with their maximum intensity values.
We utilize the automatic mixed-precision (AMP) to train signal recovery networks (our models and SSN \cite{wang2023ssn}).
To ensure concise reproduction, we follow the original source code and configurations of other learning-based methods and train these models without AMP.
We apply maximum intensity projection along $z$-axis to obtain the predicted 2D intensity maps. To compute depth maps, we apply a 10\% threshold based on the maximum intensity.

%% file: Supple/Sections/baseline_detail.tex
\section{Additional Baselines, Details and Results}
In this section, we provide details of the reconstruction procedures of the baseline methods, including configurations and sanity check of these methods.
We also provide additional baselines for more precise comparisons, which are FBP \cite{velten2012fbp} with Laplacian-of-Gaussian (LoG) and Laplacian (Lap.) filters, Phasor field with a BP solver (Phasor (BP)) \cite{liu2019phasor}, the method proposed by Mu \etal \cite{mu2022rescue} (denoted as P2R), and NLOST \cite{li2023nlost}.

\label{sec:baseline_detail}
\subsection{Baseline Details}

\parabf{FK \cite{lindell2019fk}, LCT \cite{o2018lct}, and RSD \cite{liu2020diffraction}.}
To deliver the results of inverse NLOS methods with interpolation techniques, we utilize the GPU implementation of these methods based on the code provided in \cite{chen2020lfe} and \cite{mu2022rescue}, which offers faster reconstruction time compared to the original CPU implementation.
All methods reconstruct output volumes using measurements aligned through the same procedure as our method, as described in the above.

\parabf{BP-based methods.}
We also include several BP-based solutions, which can directly reconstruct high-resolution hidden volumes from low-resolution measurements.
To report these BP-based methods, which include FBP (Lap.), FBP (LoG), Phasor (BP) \cite{liu2019phasor}, we utilize the source code provided in \cite{liu2019phasor}.
We follow the original configuration and set $\lambda = 2\Delta_p$, where $\Delta_p$ is the sampling distance, to report the results of Phasor field.

\parabf{SSCR \cite{liu2023sscr}.}
SSCR aims to address few-shot NLOS imaging through the optimization process.
We utilize the original source code provided by the authors and make adjustments to the photon counts to align with our experiment setup.
We adhere to the original configurations for Dragon, and apply the same settings to Bike.
In the case of the non-confocal Resolution instance, we adopt the configurations for the letter ``4" instance, which shares most similarities with Resolution.
Note that SSCR is often sensitive to hyperparameter configurations, and we may expect better results by heuristically tuning these parameters for each instance.

\parabf{SSN \cite{wang2023ssn}.}
SSN is the first signal recovery network that learns to predict high-resolution measurements from their low-resolution counterparts. Since the official source code of this method is not available, we reproduce SSN by following their paper. Specifically, we leverage the official source code of A2N \cite{chen2021a2n}, a baseline method of SSN designed for image super-resolution. We modify this network to use a 3D convolution kernel and a 3D attention, as described in SSN \cite{wang2023ssn}. Additionally, we follow the network configuration provided in their paper, except for the base feature dimension which is not specified in the paper. For the base feature dimension, we follow the default configuration of A2N.
We also apply AMP to train SSN in the same manner as our model.

\parabf{LFE \cite{chen2020lfe}.}
All learning-based baselines, including LFE, takes upsampled (and zero-padded for smaller aperture cases) measurements using the nearest interpolation method as inputs, which yield better performance of these models in partial sampling scenarios.
To train LFE on our synthetic dataset and sensor noise model, we incorporate the original source code of the network into our training pipeline with minimal code changes.
To ensure fair quantitative comparisons, we also modify LFE to utilize RSD as a feature propagator and eliminate a spatial downsampling operator at the beginning of the network. We observed that propagating feature volumes with excessively low spatial resolution adversely affects their final outcomes.
As we evaluate visual quality using the results of RSD with optimal measurements, we also utilize these 2D projected results as training targets. Additionally, we provide qualitative results of LFE trained using front images as training targets in \Fref{fig:supp_sanity}, to ensure that using the results of RSD as ground truth does not damage the final quality of this method.

\parabf{USM \cite{li2024usm}.}
Recently, Li \etal propose an end-to-end neural network (USM) \cite{li2024usm} that is designed to reconstruct hidden scenes from partial measurements.
USM is an extension of LFE, which places the signal recovery network in front of the volume reconstruction network (similar to LFE). Specifically, USM consists of the signal recovery network, physics-based feature propagator, volume refine module and the projection reconstruction module.
As it is an extension of LFE, same as LFE, we modify the original code of USM to utilize RSD as a feature propagator, eliminate a spatial downsampling operator at the beginning, and train this model with our noise augmentation technique. We also provide quantitative results for USM, which are reproduced with the inclusion of depth map reconstruction. 

\parabf{P2R \cite{mu2022rescue}.}
Similar with LFE, we also bring the original source code of the network into our training pipeline.
To align with the RSD configuration, we slightly modify the employed RSD propagator of this method to use the wavelength coefficient 1 and the number of cycles 5.
Same as LFE, we also omit the initial spatial downsampling operator and train this method using 2D projected results of RSD.
Since this method is not explicitly trained to render depth maps, we apply a threshold technique to obtain the depth maps, same as our method.

\parabf{NLOST \cite{li2023nlost}.}
Finally, we also utilize the original source code of NLOST and bring this into our training pipeline for reproduction.
Since this method does not involve the initial spatial downsampling, we keep the original implementation.
Same as LFE, we modify this method to employ RSD as a feature propagator, and train the network using 2D projected results of RSD.

\input{Supple/Partials/figure_sanity}
\subsection{Sanity Check}
To validate the reproduced results of the baselines, we present the results on the full measurements with $64 \times 64$ sampling resolution and the longest exposure time (180 minute for the original measurements) as a sanity check.
As shown in \Fref{fig:supp_sanity}, all baseline methods produce high-quality results with full measurements, confirming that degraded results of these baselines originate from the reduced number of samplings and scan areas in partial sampling scenarios.
The results of LFE with the initial downsampling operator (LFE (down)) lack fine details of Statue, indicating that propagating feature volumes in a low-resolution results in degradation of end-to-end learning-based methods.

\input{Supple/Partials/figure_comp_physics}
\input{Supple/Partials/figure_comp_learning}
\subsection{Additional Comparison}
For more precise comparisons, we compare our method with additional baselines.
We present quantitative results on the synthetic datasets, including additional learning-based baselines, in \Tref{table:supp_additional_comp}.
We deliver comparisons with more physics-based methods in \Fref{fig:supp_comp_physics}, and comparisons with more learning-based methods in \Fref{fig:supp_comp_learning}.
Our method outperforms all other baselines in all scenarios, demonstrating the effectiveness of the proposed denoising auto-encoder scheme with the phasor-based pipeline.

\input{Supple/Partials/table_additional_comp}
\input{Supple/Partials/table_runtime}
\parabf{Runtime and GPU Memory Comparisons.}
We compare the runtime and GPU memory usage to reconstruct hidden volumes in \Tref{table:supp_runtime_memory}.
Our method exhibits reasonable reconstruction time and memory usage compared to other learning-based methods.
While LFE \cite{chen2020lfe} reports better reconstruction time and GPU memory usage, this method fails to deliver favorable results in partial sampling scenarios, as shown in the main paper.
On the other hand, P2R \cite{mu2022rescue} and SSN \cite{wang2023ssn} suffer from the expensive computations caused by the employed volume rendering pipeline (P2R) and the attention layers (SSN).
Our method is capable of learning rich and noise-robust representations in the measurement space that are sufficient to recover missing scans from partial observations, resulting in high-quality outputs in an efficient manner.
Our model also exhibits the improved efficiency compared to USM, which employs multi-kernel feature extraction architectures as their signal recovery network.

We also compare the runtime of our method with SSCR \cite{liu2023sscr}, which is an optimization-based baseline specifically designed for partial NLOS imaging scenarios.
To ensure a fair comparison, we measure the latency of our method using a CPU, as the original implementation of SSCR is designed for CPU, using MATLAB.
Our model takes $0.7\ \textrm{s}$ for $16 \times 16$ and $0.5\ \textrm{s}$ for $8 \times 8$ measurements of the Dragon dataset \cite{lindell2019fk}. In contrast, SSCR consumes significantly more time, taking $194\ \textrm{s}$ for $16 \times 16$ and $123\ \textrm{s}$ for $8 \times 8$ measurements of the same dataset in our environment. This efficiency gap demonstrates the computational burden of optimization-based methods, which incur $O(M^2N^3)$ computations per iteration, where $M$ represents the spatial resolution of measurements and $N$ denotes the hidden volume resolution.
Please note that the runtime of SSCR could vary depending on factors such as scenes, chosen hyperparameters, and the convergence speed.
Additionally, we emphasize that our model does not involve any iterative procedures and is well-suited to leverage GPU acceleration.

%% file: Supple/Partials/figure_sanity.tex
\begin{figure}[t]
    \centering
    \includegraphics[width=1\linewidth]{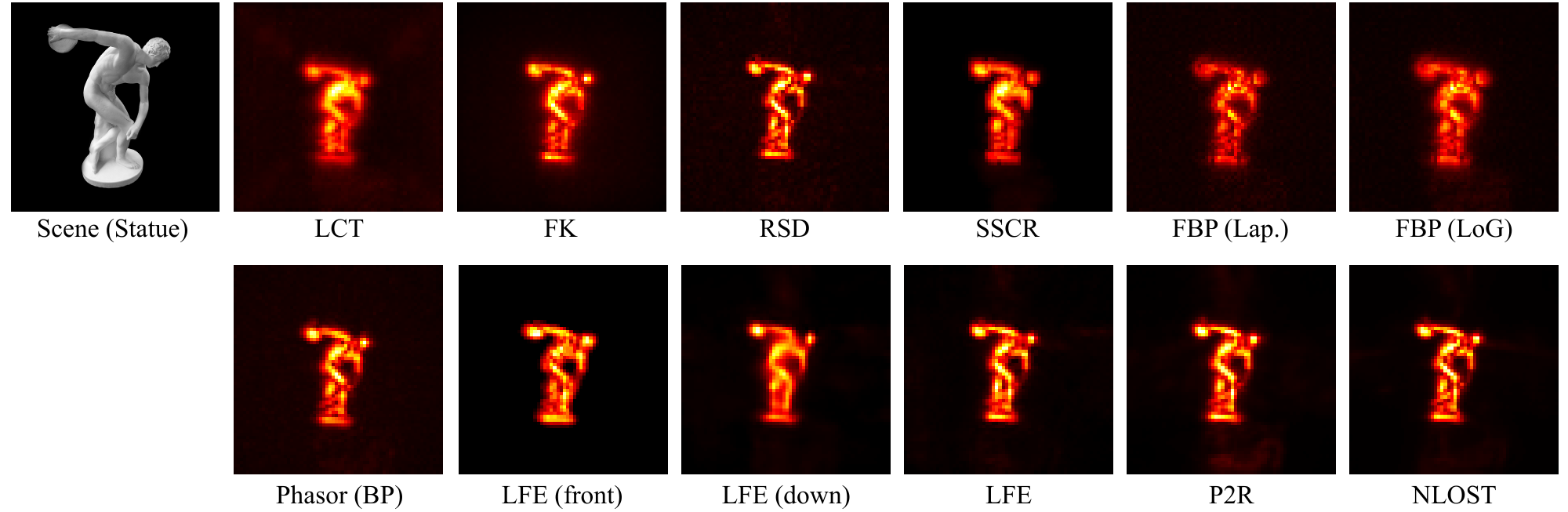}
    \caption{
    Results of baseline methods on $64 \times 64$ measurements of Statue with longest exposure time (180 minutes for the original measurements), from the Stanford dataset \cite{lindell2019fk}. LFE (front) is trained with front images generated by the renderer as ground truth intensity images, and LFE (down) includes the initial spatial downsampling operator as in the original work. All baseline methods deliver high-quality results with sufficient samplings, scan areas, and exposure time.
    }
    \label{fig:supp_sanity}
\end{figure}

%% file: Supple/Partials/figure_comp_physics.tex
\begin{figure*}[t]
    \begin{center}
        \includegraphics[width=1\linewidth]{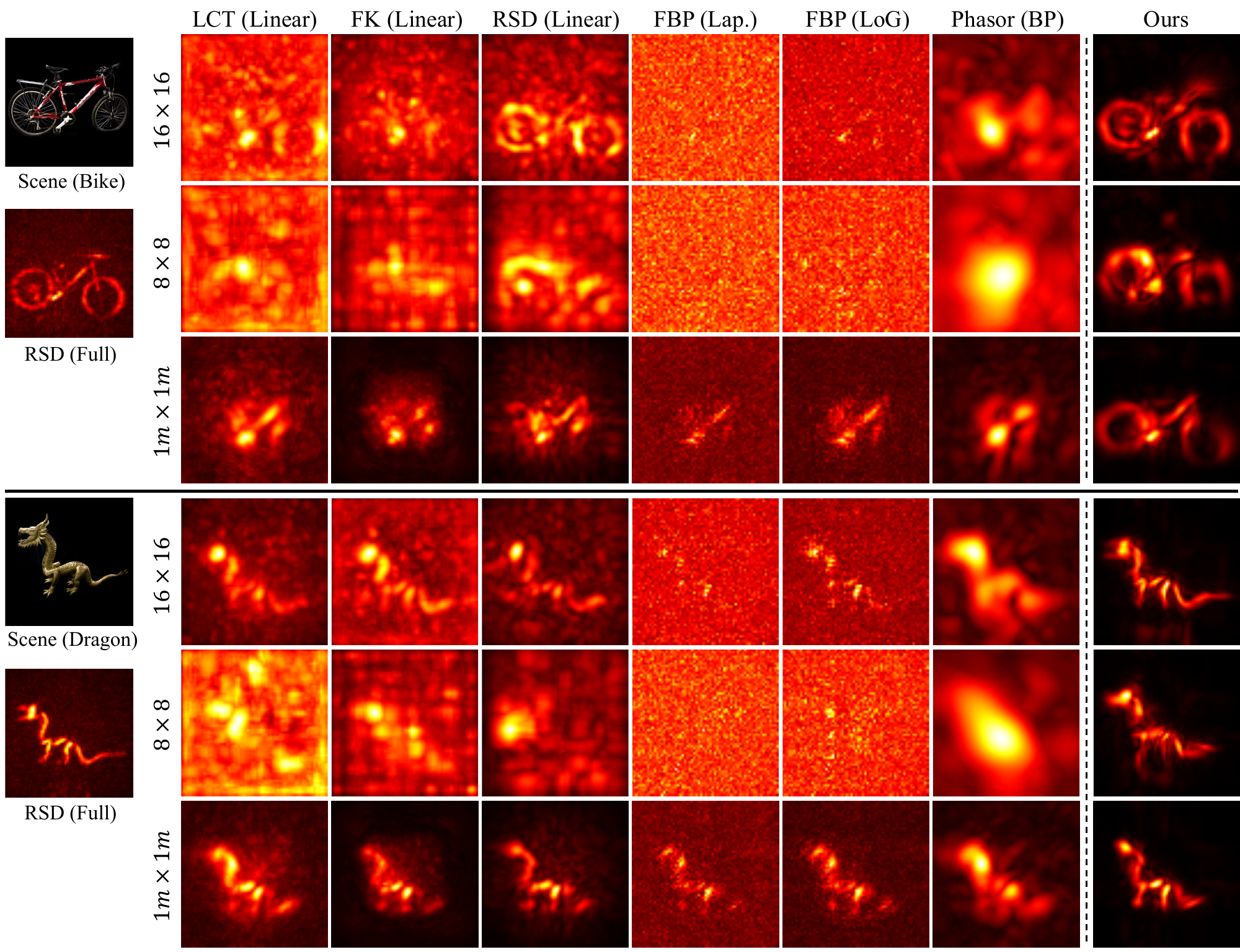}
    \end{center}
    \vspace{-10pt}
    \caption{
    Qualitative comparisons with additional inverse NLOS methods on Stanford confocal real-world dataset \cite{lindell2019fk}.
    The results of FBP \cite{velten2012fbp} with Laplacian (Lap.) and Laplacian-of-Gaussian (LoG) filters, and Phasor field \cite{liu2019phasor} with a back-projection solver with $\lambda = 4\Delta$ are included. Evaluation scenarios involve $16 \times 16$ and $8 \times 8$ sparse samplings with \sqsize{2}{m}{2}{m} apertures, and the \sqsize{1}{m}{1}{m} smaller aperture with $16 \times 16$ samplings.
    }
    \label{fig:supp_comp_physics}
\end{figure*}

%% file: Supple/Partials/figure_comp_learning.tex
\begin{figure*}[t]
    \begin{center}
        \includegraphics[width=1\linewidth]{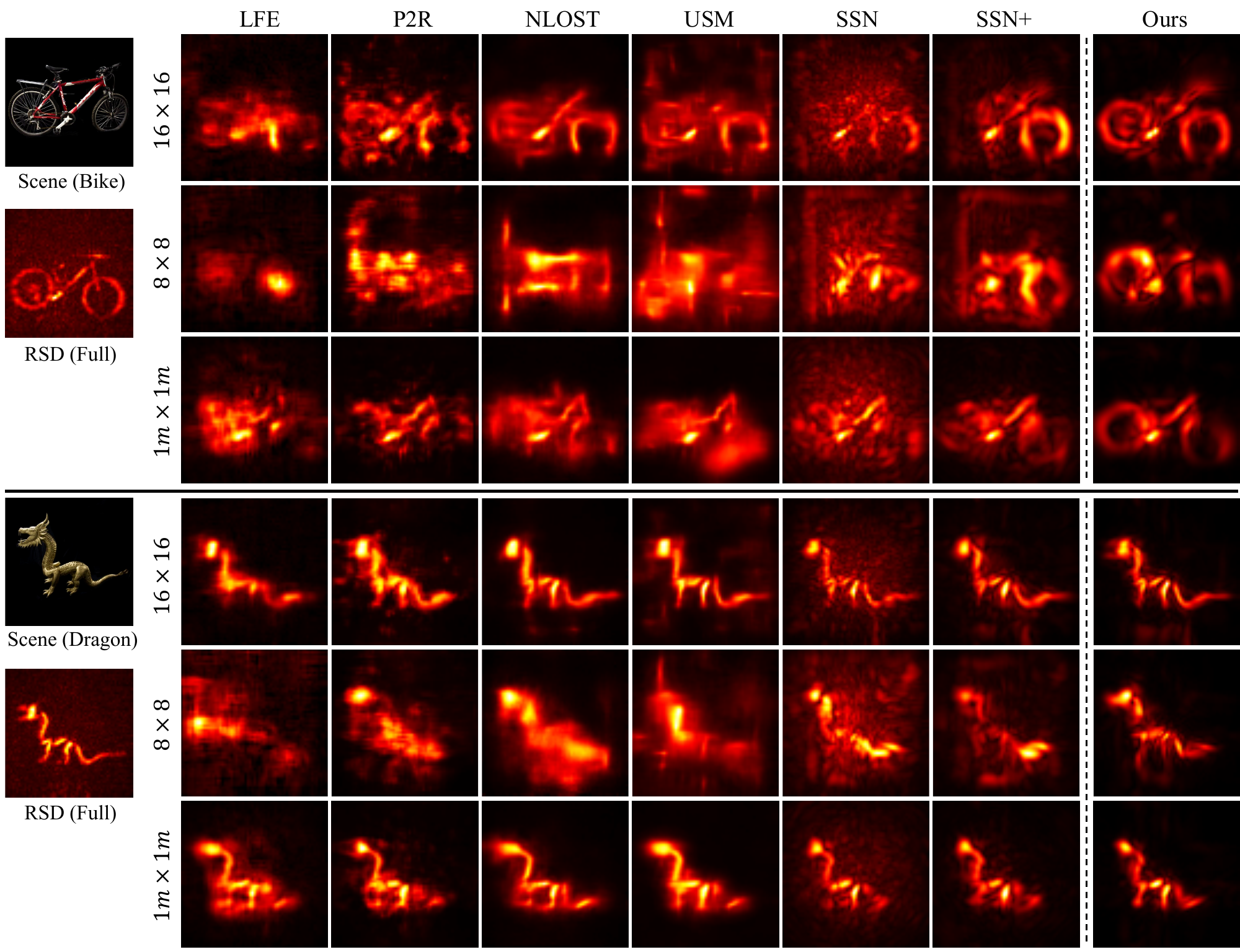}
    \end{center}
    \vspace{-10pt}
    \caption{
    Qualitative comparisons with additional learning-based baselines on Stanford confocal real-world dataset \cite{lindell2019fk}.
    Results of LFE \cite{chen2020lfe}, the model proposed by Mu \etal (denoted as P2R) \cite{mu2022rescue}, NLOST \cite{li2023nlost}, SSN \cite{wang2023ssn}, and SSN with denoising criterion (SSN+) are additionally included. Our method produces clean and detailed shapes of the objects, whereas other methods yield noisy results (P2R, NLOST, USM, SSN) or miss some details of the objects (NLOST, SSN+), \eg, the wheels of Bike, the head and the legs of Dragon. Evaluation scenarios involve $16 \times 16$ and $8 \times 8$ sparse samplings with \sqsize{2}{m}{2}{m} apertures, and the \sqsize{1}{m}{1}{m} smaller aperture with $16 \times 16$ samplings.
    }
    \label{fig:supp_comp_learning}
\end{figure*}

%% file: Supple/Partials/table_additional_comp.tex
\begin{table}[]
\setlength{\tabcolsep}{4pt}
\centering
\small
\caption{
Additional quantitative comparisons on the synthetic dataset. The method proposed by Mu \etal (denoted as P2R) \cite{mu2022rescue}, NLOST \cite{li2023nlost} are additionally included as learning-based baselines.
We also deliver the reproduced results of USM with depth reconstruction (denoted as USM (depth)).
}
\begin{tabular}{c|ccc|ccc}
\toprule
\multirow{2}{*}{Method} & \multicolumn{3}{c}{Conf-16}& \multicolumn{3}{c}{Conf-8} \\
 & PSNR & SSIM & RMSE & PSNR & SSIM & RMSE \\
\midrule
RSD\textsubscript{Nearest} & 14.85 & 0.1515 & 0.8232 & 12.65 & 0.0855 & 0.8919 \\
RSD\textsubscript{Linear} & 14.62 & 0.1631 & 0.7536 & 11.28 & 0.0760 & 0.8884 \\
LFE \cite{chen2020lfe} & 23.05 & 0.6729 & 0.2247 & 17.45 & 0.4098 & 0.3282 \\
USM \cite{li2024usm} & 29.99 & \textbf{0.8994} & - & 25.75 & 0.8235 & - \\
USM (depth) \cite{li2024usm} & 28.94 & 0.8572 & 0.1997 & 25.12 & 0.7895 & 0.1878 \\
P2R \cite{mu2022rescue} & 26.21 & 0.7742 & 0.1695 & 20.14 & 0.5830 & 0.1960 \\
NLOST \cite{li2023nlost} & 28.25 & 0.8679 & 0.1761 & 23.60 & 0.7602 & 0.2102 \\
SSN \cite{wang2023ssn} & 23.27 & 0.4506 & 0.2699 & 21.49 & 0.4426 & 0.2259 \\
Ours & \textbf{32.02} & 0.8949 & \textbf{0.0892} & \textbf{28.07} & \textbf{0.8472} & \textbf{0.0962} \\
\midrule
 & \multicolumn{3}{c}{Conf-small}& \multicolumn{3}{c}{Non-16} \\
 & PSNR & SSIM & RMSE & PSNR & SSIM & RMSE \\
\midrule
RSD\textsubscript{Nearest} & 19.73 & 0.3743 & 0.3073 & 19.67 & 0.3218 & 0.5020 \\
RSD\textsubscript{Linear} & 19.90 & 0.4664 & 0.2046 & 19.29 & 0.3224 & 0.4856 \\
LFE \cite{chen2020lfe} & 22.92 & 0.6826 & 0.2852 & 29.47 & 0.8077 & 0.2898 \\
USM \cite{li2024usm} & 26.94 & 0.8519 & - & 35.14 & 0.9313 & - \\
USM (depth) \cite{li2024usm} & 26.27 & 0.8258 & 0.2037 & 32.47 & 0.8547 & 0.2258 \\
P2R \cite{mu2022rescue} & 24.17 & 0.7194 & 0.1815 & 25.86 & 0.7675 & 0.2036 \\
NLOST \cite{li2023nlost} & 25.84 & 0.8324 & 0.1963 & 32.58 & 0.9037 & 0.1868 \\
SSN \cite{wang2023ssn} & 21.98 & 0.4629 & 0.2036 & 29.45 & 0.6367 & 0.1798 \\
Ours & \textbf{28.31} & \textbf{0.8556} & \textbf{0.0969} & \textbf{37.45} & \textbf{0.9625} & \textbf{0.1414} \\
\bottomrule

\end{tabular}
\vspace{-5pt}
\vspace{-10pt}
\label{table:supp_additional_comp}
\end{table}

%% file: Supple/Partials/table_runtime.tex
\begin{table}[]
\setlength{\tabcolsep}{7pt}
\centering
\caption{
Analysis of runtime and GPU memory usage of learning-based methods. We report the runtime and memory usage during the inference in the $16 \times 16$ sparse sampling scenario with a batch size 1. We measure the actual GPU memory usage with the ``nvidia-smi'' command.
}
\begin{tabular}{c|cc}
\toprule
Method & Runtime & GPU memory \\
\midrule
LFE \cite{chen2020lfe} & 5.7 ms & 5998 MB \\
P2R \cite{mu2022rescue} & 29.4 ms & 8500 MB \\
NLOST \cite{li2023nlost} & 26.3 ms & 6620 MB \\
SSN \cite{wang2023ssn} & 77.1 ms & 7602 MB  \\
USM \cite{li2024usm} & 70.3 ms & 9580 MB \\
Ours & 24.6 ms & 6118 MB \\
\bottomrule

\end{tabular}
\label{table:supp_runtime_memory}
\end{table}



%% file: Supple/Sections/additional_result.tex
\section{Additional Results}
In this section, we deliver supplementary experimental results to provide better understandings of our model.

\label{sec:additional_result}
\subsection{Additional Synthetic Evaluation}
\input{Supple/Partials/figure_synth_qualitative}
\input{Supple/Partials/table_quantitative_full}
\parabf{Qualitative results.}
We further provide qualitative comparisons on the synthetic dataset. As in the manuscript, we compare with RSD\textsubscript{Nearest}, RSD\textsubscript{Linear}, USM \cite{chen2020lfe}, and SSN \cite{wang2023ssn}.
As shown in \Fref{fig:supp_synth_qualitative}, the proposed LEAP delivers the best results, presenting the noise-robustness of our model and the ability to recover fine details of the objects.

\parabf{Extended quantitative results.}
We deliver the extended quantitative results on the synthetic dataset, including baselines compared in the manuscript.
We also additionally include  SSN \cite{wang2023ssn} with denoising criterion (SSN+), Ours\textsubscript{time} Ours\textsubscript{time} with denoising criterion (Ours\textsubscript{time}+) as baselines of signal recovery networks, which are compared in the ablation study of the manuscript.
We present the foreground evaluation results, which are obtained by measuring the scores only in the foreground regions.
We determine the foreground masks using ground truth depth maps. To ensure sufficient coverage of foreground objects, we slightly increase the foreground regions by applying average pooling to the foreground masks with a kernel size of 5.

\Tref{table:supp_quantitative_full} reports the extended quantitative results on the synthetic dataset.
Our method outperforms all other methods, including several signal recovery networks, in all metrics except SSIM in confocal $16 \times 8$ and $8 \times 8$ sparse sampling scenarios.
While Ours\textsubscript{time}+ often delivers slightly better results in SSIM measured from the entire regions in confocal $16 \times 16$ and $8\ times 8$ sparse sampling scenarios, our method with phasor-based frequency management scheme excels in reconstructing finer details of objects, as evidenced by scores measured specifically in the foreground regions.
Meanwhile, the performance gap of the RSD with interpolation methods between foreground and all regions are more significant in the smaller aperture scenario.
This can be attributed to the zero-padding applied for these methods to match the aperture size with the reconstruction resolution.
The padded histograms with zero photon counts suppress the effects of the background noise, resulting in better performance of the interpolation methods particularly in the background regions.


\input{Supple/Partials/figure_exposure180}
\input{Supple/Partials/figure_exposure30}
\subsection{Evaluation With Various Exposure time}
We evaluate the noise robustness of the methods by comparing the results with shorter and longer exposure time.
\Fref{fig:supp_exposure180} delivers the results on Stanford confocal real-world dataset \cite{lindell2019fk} with longer exposure time, corresponding to a 180 minute exposure time for the original measurements. This results in $42.2\ \textrm{s}$ scanning time for $16 \times 16$ samplings, and $42.2\ \textrm{s}$ scanning time for $8 \times 8$ samplings.
We also deliver the results with a shorter exposure time in \Fref{fig:supp_exposure30}, which corresponds to a 30 minute exposure time for the original measurements. In this case, $16 \times 16$ samplings require $7.0\ \textrm{s}$ scanning time and $8 \times 8$ samplings require $1.8\ \textrm{s}$ scanning time.
While longer exposure time leads to the cleaner results, our method still produces reasonable outputs with shorter exposure time. In addition, we observe that $16 \times 16$ samplings with shorter exposure time offer better results than $8 \times 8$ samplings with longer exposure time, which can be attributed to the difficulty of inferring missing pixels between larger sampling distances. We expect stronger generative priors, such as generative diffusion models \cite{ho2020ddpm, song2020score}, could be a solution for addressing this difficulty.

\input{Supple/Partials/table_ablation_all}
\input{Supple/Partials/figure_incorp_others}
\subsection{Incorporating With Other NLOS Methods}
\label{sec:incorporating_others}
To incorporate with other inverse NLOS methods, \eg LCT \cite{o2018lct} and FK \cite{lindell2019fk}, we can consider the intermediate predictions of our method as frequency-filtered measurements. After obtaining the intermediate predictions $\hat{H}(\mathbf{x_p \rightarrow x_c}, \Omega)$, we convert them into the time domain predictions $\hat{H}(\mathbf{x_p \rightarrow x_c}, t)$ by applying the inverse Fourier transform. The resulting measurements can be viewed as measurements processed by an appropriate band-pass filter.
Then the inverse NLOS methods can be applied to these time domain predictions to reconstruct hidden scenes.
We deliver the results of our method incorporating with two representative inverse methods, namely LCT and FK, in \Fref{fig:supp_incorp_others}. We also report the results of SSN \cite{wang2023ssn} and SSN with denoising criterion (SSN+), employing LCT and FK as propagators, for comparisons.

\subsection{Additional Ablation Study}
We further provide additional ablation results for a deeper understanding of the proposed method.
Consistent with the manuscript, experiments are carried out in $16 \times 16$ sparse sampling scenarios, encompassing both confocal and non-confocal imaging systems.

\parabf{Loss function.}
While most of the previous learning-based methods \cite{chen2020lfe, mu2022rescue, li2023nlost, wang2023ssn} employ the mean-square-error (MSE) as their training objective, we empirically observe that the L1 loss yields slightly better performance for our model.
The ablation results on the training objective are presented in the top of \Tref{table:supp_ablation_all}.
As can be seen, employing L1 loss leads to a slight improvement in the performance of our model, while MSE loss generally performs better for SSN and SSN with denoising criterion (SSN+). Consequently, we adhere to the original configuration of SSN and present the results of SSN with the MSE loss in all experiments conducted in both our main paper and the supplementary material.

\parabf{The number of input wavelengths.}
We further analyze the effects of the number of the wavelengths for the input phasor field convolution.
We change the wavelength coefficients to $\{1.5\}$ (num. $\lambda = 1$), $\{1.25, 1.5, 2.0\}$ (num. $\lambda = 3$), $\{1.0, 1.25, 1.5, 2.0, 2.5\}$ (num. $\lambda = 5$), and our model with the original configuration using $\{0.8, 0.9, 1.0, 1.25, 1.5, 2.0, 2.5\}$ (num. $\lambda = 7$, Ours).
The quantitative ablation results on the number of input wavelengths are provided in the second group of \Tref{table:supp_ablation_all}. It can be observed that increasing the number of input wavelengths brings gradual increase of the performance of our model.
These results indicate that increasing the number of input frequency ranges near the frequency range of interests leads to the performance improvement in both visual quality (PSNR) and accuracy of reconstructed geometry (RMSE).

\parabf{Noise with multiple exposure levels.}
To test the effects training with multiple exposure levels, we conduct the ablation on our noise augmentation scheme.
We compare with three noise models: noise models with (1) a fixed exposure level $c = 1$, (2) a fixed exposure level $c = 0.5$, and a fixed exposure level $c = 1$. As reported in the third group of \Tref{table:supp_ablation_all}, the model trained with a exposure level randomly sampled from the range $[0.1, 1.0]$ yields best performance. The model trained with a shortest exposure level $c = 0.1$ results in the notable degradation of the performance, indicating that only using an extreme noise model poses difficulties in the entire training pipeline.
These noise models are applied only in the training and all models are validated using the same sensor model $c \in [0.1, 1.0]$.

\input{Supple/Partials/figure_ablation_frequency}
\parabf{Phasor domain processing.}
In addition to Fig. 7 of the main paper, to further examine the effects of the frequency filtering and our phasor domain processing, we deliver the results of the ``all-pass'' model, which is a phasor-based enhancement network but uses all frequencies.
As reported in the bottom of \Tref{table:supp_ablation_all}, the proposed model (Ours) achieves the highest performance.
Interestingly, the ``low-pass'' model shows degraded results compared to the ``all-pass'' model, which we attribute to the higher number of high-frequency components compared to the low-frequency range.
This causes the ``low-pass'' model to concentrate on a smaller number of frequency components, amplifying the effects of spurious low frequency signals.
Such phenomenon can also be observed in the results of the ``all-pass model (equal)'', where the effects of all frequency ranges (ranges A, B, C in Fig. 2 of the main paper) become equal regardless of the number of frequency components due to the supervision with rescaled loss weights.
While our phasor domain processing also contribute to the improvement, the proposed frequency filtering clearly presents its effectiveness, particularly in capturing detailed shapes of the object, as clearly highlighted in \Fref{fig:supp_ablation_frequency}.

\input{Supple/Partials/figure_depth}
\subsection{Depth Map Visualization}
We deliver the results of reconstructed depth maps in \Fref{fig:supp_depth}. Depth maps of all models except LFE \cite{chen2020lfe} are obtained using a 10\% threshold of the maximum intensity, since LFE is directly learned to predict the 2D depth maps.
As can be seen, the proposed model produces clean and accurate depth maps compared to other baselines.

\input{Supple/Partials/figure_confocal_other_instances}
\input{Supple/Partials/figure_statue_exposure}
\input{Supple/Partials/figure_teaser_exposure}
\input{Supple/Partials/figure_nonconfocal_other_instances}
\subsection{Results on Other Instances}
\label{sec:other_instances}
We additionally present results on various real-world instances with both confocal and non-confocal systems.
We deliver the results on Statue, Teaser instances from Stanford confocal real-world dataset, in the $16 \times 16$, $8 \times 8$ sparse scanning scenarios and the \sqsize{1}{m}{1}{m} smaller aperture scenario.
Since measurements of these instances have higher photon counts than Bike and Dragon, we compare the results on measurements with a shorter exposure time ($9\ \textrm{ms}$ per pixel).
As reported in \Fref{fig:supp_confocal_other_instances}, our model outperforms all other baseline methods, producing cleanest results with fine details of the scenes.
We also deliver the results of learning-based methods with a shorter ($9\ \textrm{ms}$ per pixel) and a longer ($27\ \textrm{ms}$ per pixel) exposure times in \Fref{fig:supp_statue_exposure} and \Fref{fig:supp_teaser_exposure}.

Furthermore, we deliver the results on additional instances, namely ``4'', ``NLOS'', and Shelf instances from the non-confocal real-world dataset \cite{liu2020diffraction}.
To better assess the performance, since these measurements are captured with sufficiently long exposure time, we compare the results in the more challenging $8 \times 8$ sparse sampling scenario.
As shown in \Fref{fig:supp_nonconfocal_other_instances}, our method yields high-quality results, reconstructing clear shapes of the instances.
On the other hand, all other baseline methods produce inaccurate shapes of the objects. As discussed in the manuscript, the \naive application of the denoising criterion to SSN (SSN+) does not achieve meaningful improvement, producing degraded and over-smoothed results in many cases.

%% file: Supple/Partials/figure_synth_qualitative.tex
\begin{figure*}[t]
\begin{center}
\includegraphics[width=0.83\linewidth]{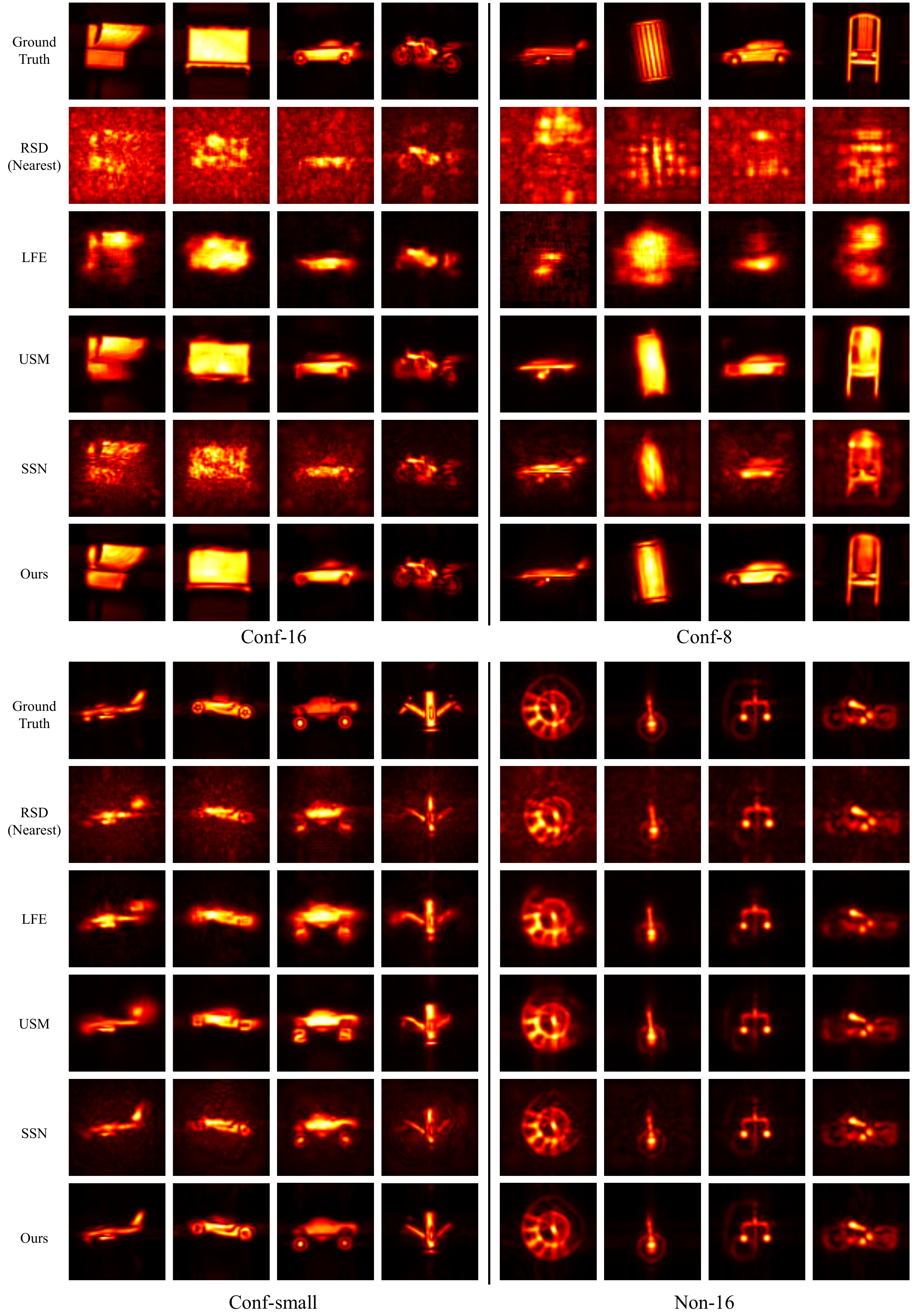}
\end{center}
   \vspace{-10pt}
   \caption{
   Qualitative results on the synthetic dataset. Results on all scenarios are reported, which include confocal $16 \times 16$ sparse samplings (Conf-16), confocal $8 \times 8$ sparse samplings (Conf-8), confocal \sqsize{1}{m}{1}{m} smaller aperture scanning with $16 \times 16$ samplings (Conf-small), and non-confocal $16 \times 16$ sparse samplings (Non-16).
   }
   \label{fig:supp_synth_qualitative}
\end{figure*}

%% file: Supple/Partials/table_quantitative_full.tex
\begin{table*}[t!]
\setlength{\tabcolsep}{4pt}
\centering
\small
\caption{Extended quantitative results on the synthetic dataset. Scores within foreground regions, obtained using ground truth foreground masks, are additionally included. We also deliver the results of signal recovery network baselines, which are SSN \cite{wang2023ssn} with denoising criterion (SSN+), our enhancement network in the time domain (Ours\textsubscript{time}), and Ours\textsubscript{time} with denoising criterion (Ours\textsubscript{time}+).
}
\resizebox{\columnwidth}{!}{
\begin{tabular}{c|ccc|ccc|ccc|ccc}
\toprule
\multirow{3}{*}{Method} & \multicolumn{6}{c|}{Conf-16} & \multicolumn{6}{c}{Conf-8} \\
\cline{2-13}
& \multicolumn{3}{c|}{all} & \multicolumn{3}{c|}{foreground} & 
\multicolumn{3}{c|}{all} & \multicolumn{3}{c}{foreground} \\
& PSNR$\uparrow$ & SSIM$\uparrow$ & RMSE$\downarrow$ & PSNR$\uparrow$ & SSIM$\uparrow$ & RMSE$\downarrow$
& PSNR$\uparrow$ & SSIM$\uparrow$ & RMSE$\downarrow$ & PSNR$\uparrow$ & SSIM$\uparrow$ & RMSE$\downarrow$ \\
\hline

RSD\textsubscript{Nearest} & 14.85 & 0.1515 & 0.8232 & 12.70 & 0.2512 & 0.6538 & 12.65 & 0.0855 & 0.8919 & 10.18 & 0.0762 & 0.7295 \\
RSD\textsubscript{Linear} & 14.62 & 0.1631 & 0.7536 & 12.52 & 0.2635 & 0.6448 & 11.28 & 0.0760 & 0.8884 & 9.48 & 0.0726 & 0.7278 \\
LFE \cite{chen2020lfe} & 23.05 & 0.6729 & 0.2247 & 17.46 & 0.6137 & 0.5879 & 17.45 & 0.4098 & 0.3282 & 12.49 & 0.3221 & 0.7271 \\
USM \cite{li2024usm} & 29.99 & 0.8994 & - & 24.17 & 0.8583 &- & 25.75 & 0.8235 & - & 19.89 & 0.7285 & - \\
SSN \cite{wang2023ssn} & 23.27 & 0.4506 & 0.2699 & 20.89 & 0.6834 & 0.4479 & 21.49 & 0.4426 & 0.2259 & 17.27 & 0.5526 & 0.4504 \\
SSN+ & 29.55 & 0.8907 & 0.0949 & 23.61 & 0.8359 & 0.3476 & 25.77 & 0.8290 & 0.1043 & 19.79 & 0.7046 & 0.3820 \\
Ours\textsubscript{time} & 23.85 & 0.4908 & 0.2352 & 21.48 & 0.7123 & 0.4117 & 22.39 & 0.5300 & 0.1454 & 17.40 & 0.5727 & 0.4143 \\
Ours\textsubscript{time}+ & 30.69 & \textbf{0.9057} & 0.0924 & 24.75 & 0.8634 & 0.3377 & 27.19 & \textbf{0.8566} & 0.0993 & 21.21 & 0.7607 & 0.3646 \\
\hline
Ours & \textbf{32.02} & 0.8949 & \textbf{0.0892} & \textbf{26.1}5 & \textbf{0.8886} & \textbf{0.3258} & 
\textbf{28.07} & 0.8472 & \textbf{0.0962} & \textbf{22.11} & \textbf{0.7863} & \textbf{0.3524} \\

\midrule
\multirow{3}{*}{Method} & \multicolumn{6}{c|}{Conf-small} & \multicolumn{6}{c}{Non-16} \\
\cline{2-13}
& \multicolumn{3}{c|}{all} & \multicolumn{3}{c|}{foreground} & 
\multicolumn{3}{c|}{all} & \multicolumn{3}{c}{foreground} \\
& PSNR$\uparrow$ & SSIM$\uparrow$ & RMSE$\downarrow$ & PSNR$\uparrow$ & SSIM$\uparrow$ & RMSE$\downarrow$
& PSNR$\uparrow$ & SSIM$\uparrow$ & RMSE$\downarrow$ & PSNR$\uparrow$ & SSIM$\uparrow$ & RMSE$\downarrow$ \\
\hline

RSD\textsubscript{Nearest} & 19.73 & 0.3743 & 0.3073 & 15.22 & 0.4515 & 0.5263 & 19.67 & 0.3218 & 0.5020 & 16.67 & 0.4345 & 0.6277 \\
RSD\textsubscript{Linear} & 19.90 & 0.4664 & 0.2046 & 14.82 & 0.4571 & 0.5033 & 19.29 & 0.3224 & 0.4856 & 16.24 & 0.4361 & 0.6378 \\
LFE \cite{chen2020lfe} & 22.92 & 0.6826 & 0.2852 & 17.31 & 0.6158 & 0.5840 & 29.47 & 0.8077 & 0.2898 & 24.18 & 0.8097 & 0.6156 \\
USM \cite{li2024usm} & 26.94 & 0.8519 & - & 21.09 & 0.7820 & - & 35.14 & 0.9313 & - & 29.59 & 0.9109 & -  \\
SSN \cite{wang2023ssn} & 21.98 & 0.4629 & 0.2036 & 17.61 & 0.5694 & 0.4754 & 29.45 & 0.6367 & 0.1798 & 27.62 & 0.8155 & 0.5339 \\
SSN+ & 25.52 & 0.8140 & 0.1066 & 19.58 & 0.7065 & 0.3902 & 35.47 & 0.9466 & 0.1435 & 29.70 & 0.9113 & 0.5159 \\
Ours\textsubscript{time} & 22.64 & 0.5200 & 0.1582 & 17.87 & 0.5956 & 0.4382 & 29.89 & 0.6557 & 0.1736 & 28.07 & 0.8302 & 0.5302 \\
Ours\textsubscript{time}+ & 27.24 & 0.8388 & 0.1004 & 21.32 & 0.7650 & 0.3681 & 36.36 & 0.9544 & 0.1429 & 30.57 & 0.9236 & 0.5140 \\
\hline
Ours & \textbf{28.31} & \textbf{0.8556} & \textbf{0.0969} & \textbf{22.35} & \textbf{0.7982} & \textbf{0.3555} & \textbf{37.45} & \textbf{0.9625} & \textbf{0.1414} & \textbf{31.65} & \textbf{0.9386} & \textbf{0.5091} \\
\bottomrule

\end{tabular}
}
\label{table:supp_quantitative_full}
\end{table*}

%% file: Supple/Partials/figure_exposure180.tex
\begin{figure*}[t]
    \begin{center}
        \includegraphics[width=1\linewidth]{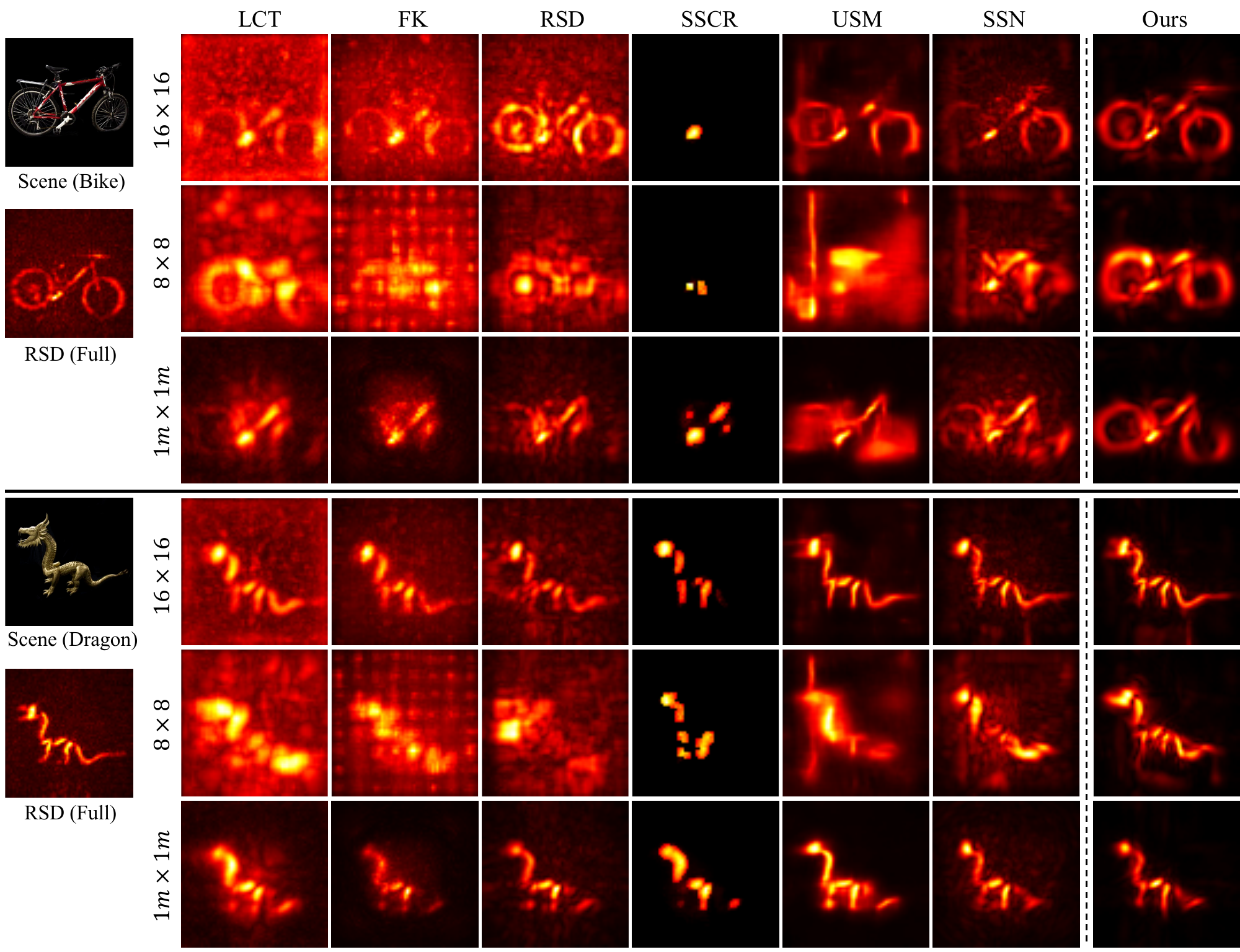}
    \end{center}
    \vspace{-10pt}
    \caption{
    Qualitative results on Bike, Dragon from Stanford real-world dataset \cite{lindell2019fk}. Results on the measurements with longer exposure time ($165\ \textrm{ms}$ per pixel, corresponding to a 180 minute exposure time for the original measurements) are reported. Evaluation scenarios involve $16 \times 16$ and $8 \times 8$ sparse samplings with \sqsize{2}{m}{2}{m} apertures, and the \sqsize{1}{m}{1}{m} smaller aperture with $16 \times 16$ samplings.
    }
    \label{fig:supp_exposure180}
\end{figure*}

%% file: Supple/Partials/figure_exposure30.tex
\begin{figure*}[t]
    \begin{center}
        \includegraphics[width=1\linewidth]{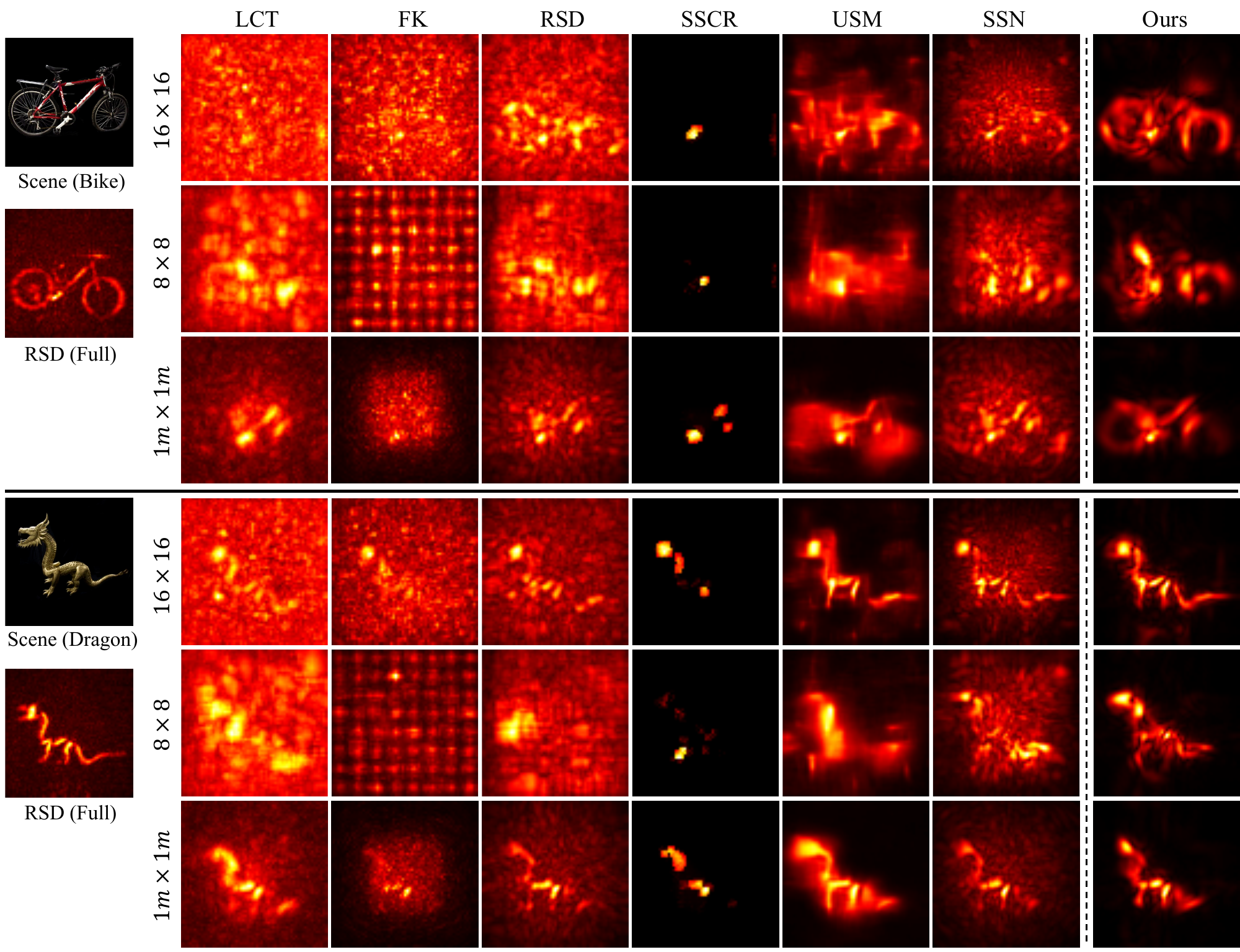}
    \end{center}
    \vspace{-10pt}
    \caption{
    Qualitative results on Bike, Dragon from Stanford real-world dataset \cite{lindell2019fk}. Results on the measurements with shorter exposure time ($27\ \textrm{ms}$ per pixel, corresponding to a 30 minute exposure time for the original measurements) are reported. Evaluation scenarios involve $16 \times 16$ and $8 \times 8$ sparse samplings with \sqsize{2}{m}{2}{m} apertures, and the \sqsize{1}{m}{1}{m} smaller aperture with $16 \times 16$ samplings.
    }
    \label{fig:supp_exposure30}
\end{figure*}

%% file: Supple/Partials/table_ablation_all.tex
\begin{table}[]
\setlength{\tabcolsep}{4pt}
\centering
\small
\caption{
Quantitative ablation results on the synthetic dataset. From top-to-bottom: Ablation results on the loss function. Ablation results on the number of wavelengths in the input phasor field convolution. Ablation results on the multiple exposure levels. Extended ablation results on the frequency filtering. The ``all-pass model (equal)'' is trained with rescaled loss weights to make effects of high and low frequency ranges equal, regardless of the number of frequency components.
}
\begin{tabular}{c|cc|cc}
\toprule
\multirow{2}{*}{Method} & \multicolumn{2}{c}{Conf-16}& \multicolumn{2}{c}{Non-16} \\
 & PSNR$\uparrow$ & RMSE$\downarrow$ & PSNR$\uparrow$ & RMSE$\downarrow$ \\
\midrule
SSN (MSE) & 23.27 & 0.2699 & 29.45 & 0.1798 \\
SSN (L1) & 23.82 & 0.2328 & 21.05 & 0.2834 \\
SSN+ (MSE) & 29.55 & 0.0949 & 35.47 & 0.1435 \\
SSN+ (L1) & 28.94 & 0.0950 & 35.09 & 0.1435 \\
Ours (MSE) & 31.84 & 0.0896 & 37.32 & 0.1415 \\
Ours (L1) & \textbf{32.02} & \textbf{0.0892} & \textbf{37.45} & \textbf{0.1414} \\
\midrule
num. $\lambda = 1$ & 30.83 & 0.0919 & 36.55 & 0.1424 \\
num. $\lambda = 3$ & 31.55 & 0.0901 & 37.04 & 0.1418 \\
num. $\lambda = 5$ & 31.82 & 0.0894 & 37.27 & 0.1416 \\
num. $\lambda = 7$ & \textbf{32.02} & \textbf{0.0892} & \textbf{37.45} & \textbf{0.1414} \\
\midrule
noise level $c = 0.1$ & 30.57 & 0.0907 & 35.62 & 0.1422 \\
noise level $c = 0.5$ & 31.92 & 0.0894 & 37.39 & 0.1415 \\
noise level $c = 1.0$ & 31.73 & 0.0894 & 37.29 & 0.1417 \\
noise level $c \in [0.1, 1.0]$ & \textbf{32.02} & \textbf{0.0892} & \textbf{37.45} & \textbf{0.1414} \\
\midrule
all-pass model & 31.38 & 0.0908 & 36.97 & 0.1421 \\
all-pass model (equal) & 31.16 & 0.0909 & 36.74 & 0.1421 \\
low-pass model & 31.23 & 0.0903 & 36.72 & 0.1420 \\
high-pass model & 31.54 & 0.0909 & 37.09 & 0.1422 \\
Ours & \textbf{32.02} & \textbf{0.0892} & \textbf{37.45} & \textbf{0.1414} \\
\bottomrule

\end{tabular}
\label{table:supp_ablation_all}
\end{table}

%% file: Supple/Partials/figure_incorp_others.tex
\begin{figure*}[t]
    \begin{center}
        \includegraphics[width=1\linewidth]{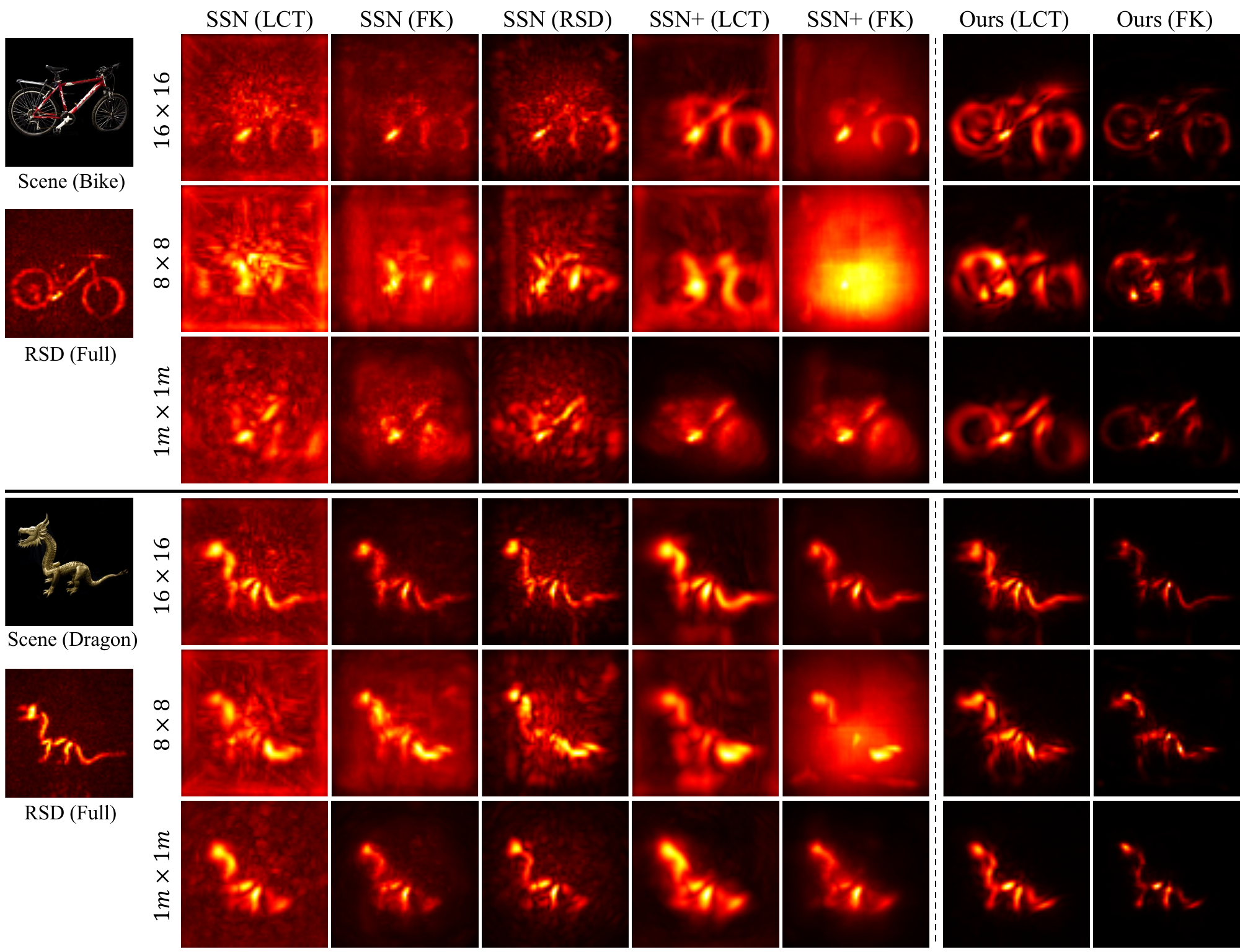}
    \end{center}
    \vspace{-15pt}
    \caption{
    Qualitative results of signal recovery networks incorporated with various inverse NLOS methods in the time domain, on Bike, Dragon measurements from Stanford confocal real-world dataset \cite{lindell2019fk}. These include SSN \cite{wang2023ssn}, SSN with denoising criterion (SSN+), and our model producing frequency-filtered measurements in the time domain. We adopt two representative inverse NLOS methods, LCT \cite{o2018lct} and FK \cite{lindell2019fk}. Evaluation scenarios involve $16 \times 16$ and $8 \times 8$ sparse samplings with \sqsize{2}{m}{2}{m} apertures, and the \sqsize{1}{m}{1}{m} smaller aperture with $16 \times 16$ samplings.
    }
    \label{fig:supp_incorp_others}
    \vspace{-10pt}
\end{figure*}

%% file: Supple/Partials/figure_ablation_frequency.tex
\begin{figure}
\centering
    \includegraphics[width=\linewidth]{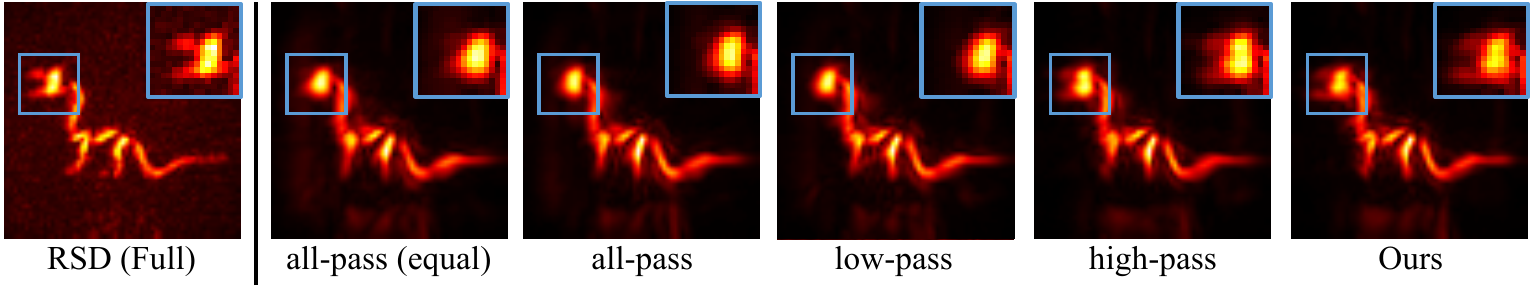}
    \caption{
    Qualitative ablation results on the frequency filtering with the all-pass model.
    The ``all-pass model (equal)'' is trained with rescaled loss weights to make effects of high and low frequency ranges equal, regardless of the number of frequency components.
    Models without low-frequency filtering fail to reconstruct details of the objects (highlighted in the blue boxes).
    }
    \label{fig:supp_ablation_frequency}
\end{figure}

%% file: Supple/Partials/figure_depth.tex
\begin{figure*}[t]
    \begin{center}
        \includegraphics[width=1\linewidth]{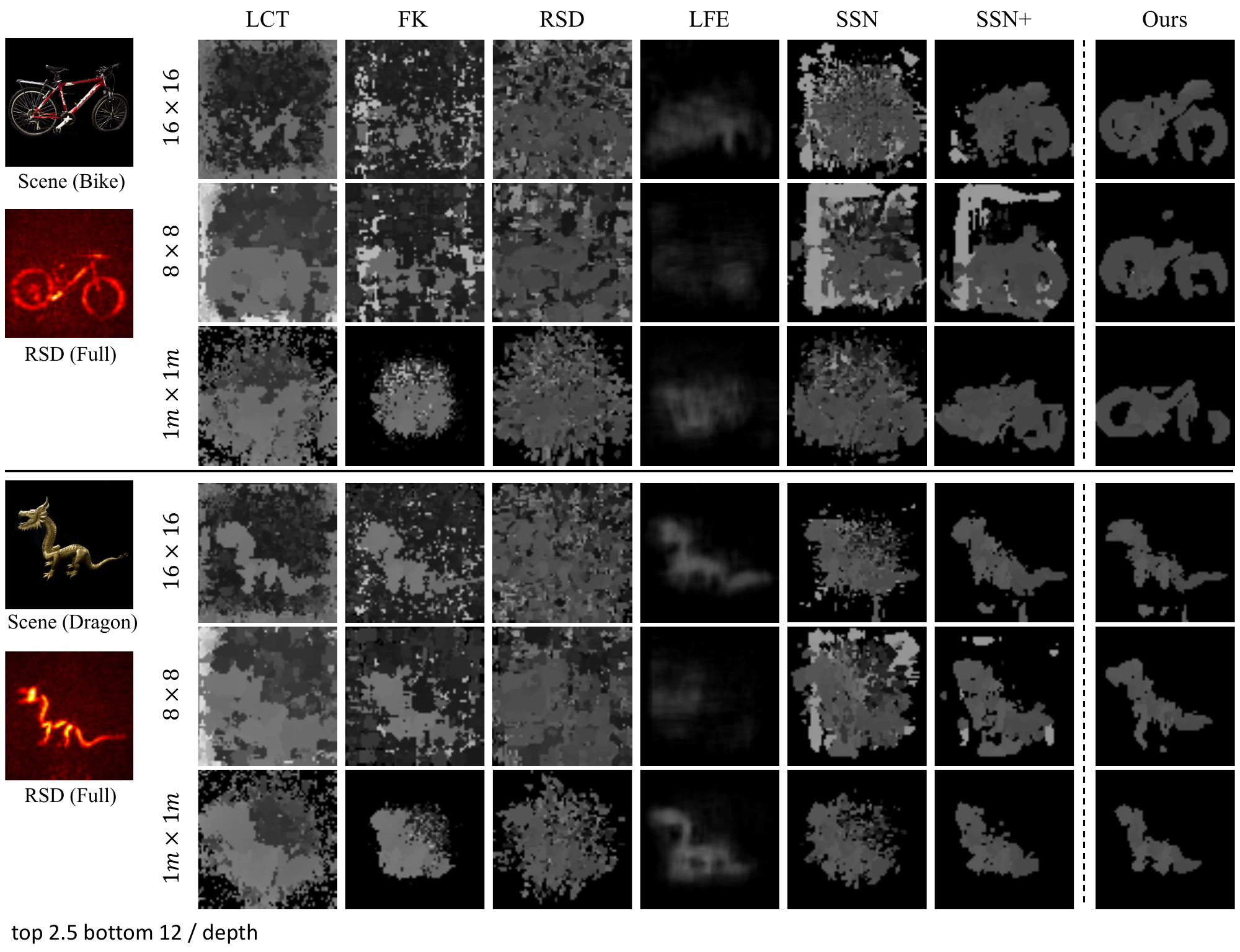}
    \end{center}
    \vspace{-10pt}
    \caption{
    Reconstructed depth maps on the Stanford confocal real-world dataset \cite{lindell2019fk}. Evaluation scenarios involve $16 \times 16$ and $8 \times 8$ sparse samplings with \sqsize{2}{m}{2}{m} apertures, and the \sqsize{1}{m}{1}{m} smaller aperture with $16 \times 16$ samplings. Results of USM \cite{li2024usm} are omitted as its original version does not incorporate depth map reconstruction.
    }
    \label{fig:supp_depth}
\end{figure*}

%% file: Supple/Partials/figure_confocal_other_instances.tex
\begin{figure*}[t]
    \begin{center}
        \includegraphics[width=1\linewidth]{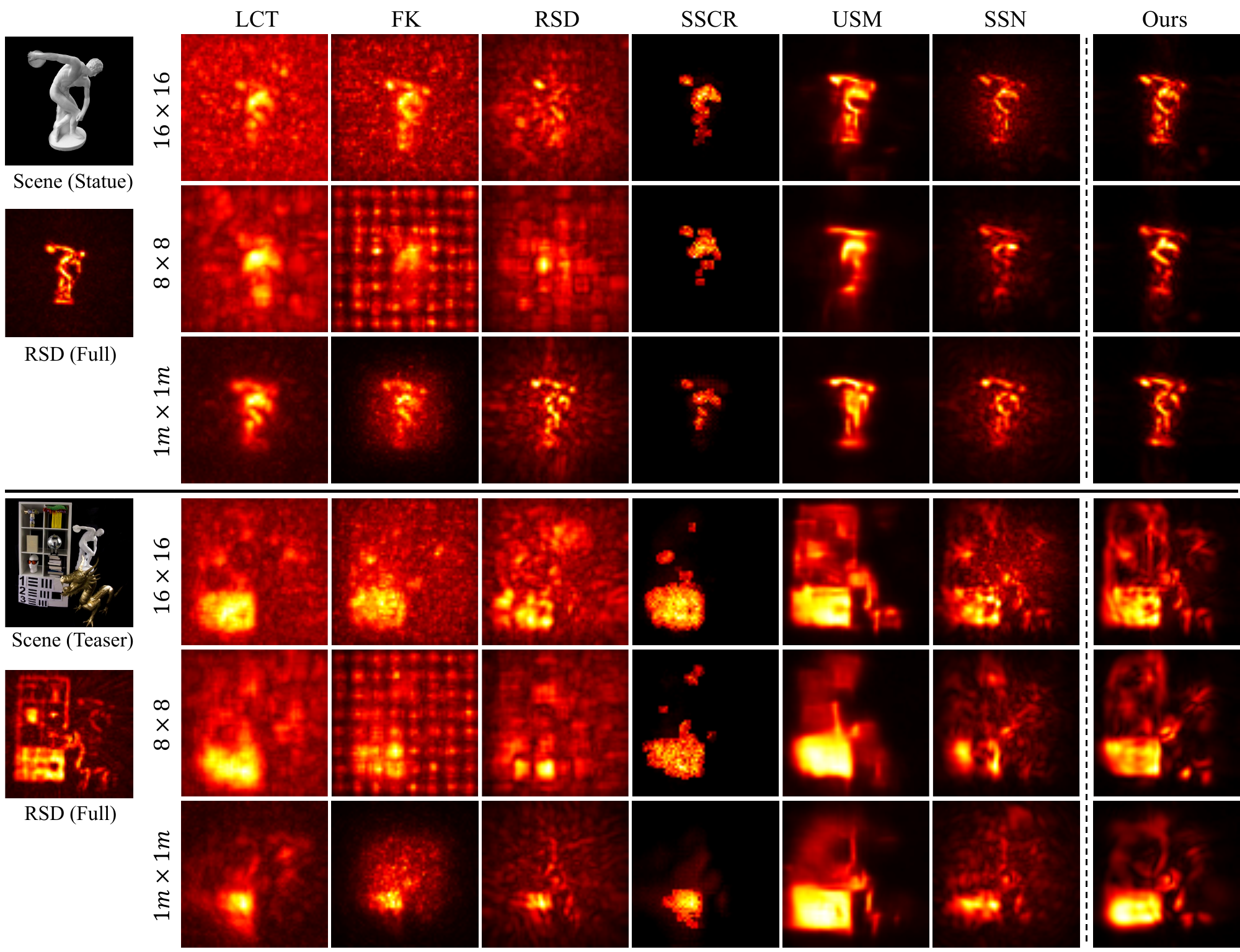}
    \end{center}
    \vspace{-10pt}
    \caption{
    Qualitative results on Statue, Teaser from the Stanford real-world dataset \cite{lindell2019fk}. We deliver the results on measurements with a $9\ \textrm{ms}$ exposure time per pixel. Evaluation scenarios involve $16 \times 16$ and $8 \times 8$ sparse samplings with \sqsize{2}{m}{2}{m} apertures, and the \sqsize{1}{m}{1}{m} smaller aperture with $16 \times 16$ samplings.
    }
    \label{fig:supp_confocal_other_instances}
    \vspace{-10pt}
\end{figure*}

%% file: Supple/Partials/figure_statue_exposure.tex
\begin{figure}[t]
    \centering
    \includegraphics[width=1\linewidth]{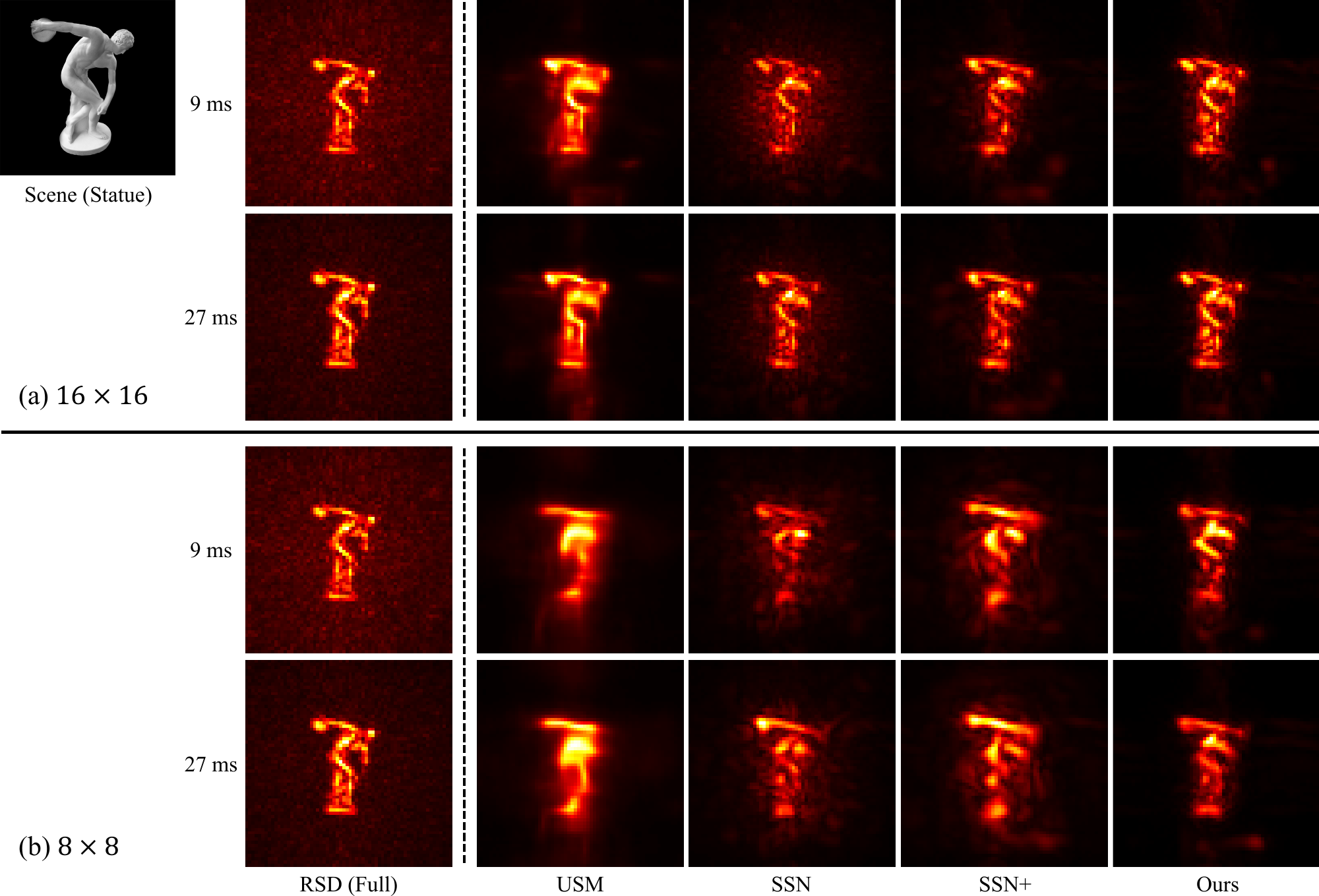}
    \vspace{-10pt}
    \caption{
    Results on the confocal measurements of Statue \cite{lindell2019fk}, with a shorter ($9\ \textrm{ms}$ per pixel) and a longer ($27\ \textrm{ms}$) exposure time per pixel. Our method reconstructs cleanest shapes of the object with details, whereas other methods produce noisy outputs (SSN), or fail to reconstruct the right arm or the legs of Statue (SSN+, USM). Evaluation scenarios involve $16 \times 16$ and $8 \times 8$ sparse samplings with \sqsize{2}{m}{2}{m} apertures.
    }
    \label{fig:supp_statue_exposure}
\end{figure}

%% file: Supple/Partials/figure_teaser_exposure.tex
\begin{figure}[t]
    \centering
    \includegraphics[width=1\linewidth]{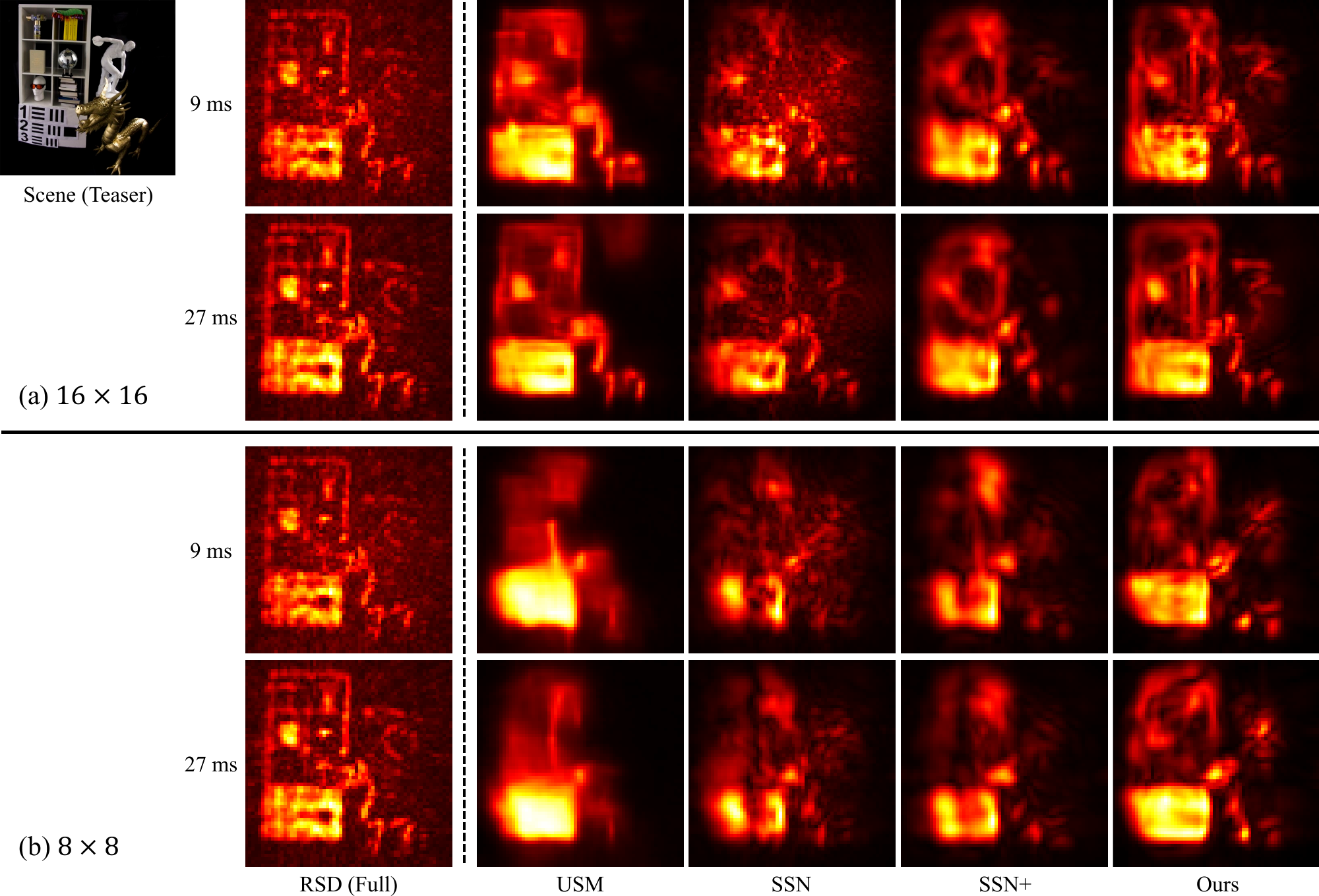}
    \caption{
    Results on the confocal measurements of Teaser \cite{lindell2019fk}, with a shorter ($9\ \textrm{ms}$ per pixel) and a longer ($27\ \textrm{ms}$) exposure time per pixel. Our method successfully reconstructs clean shapes of the scene with both exposure times, while other methods yield noisy outputs (SSN), or fail to reveal the statue (USM, SSN+). Evaluation scenarios involve $16 \times 16$ and $8 \times 8$ sparse samplings with \sqsize{2}{m}{2}{m} apertures.
    }
    \label{fig:supp_teaser_exposure}
\end{figure}

%% file: Supple/Partials/figure_nonconfocal_other_instances.tex
\begin{figure*}[t]
    \begin{center}
        \includegraphics[width=1\linewidth]{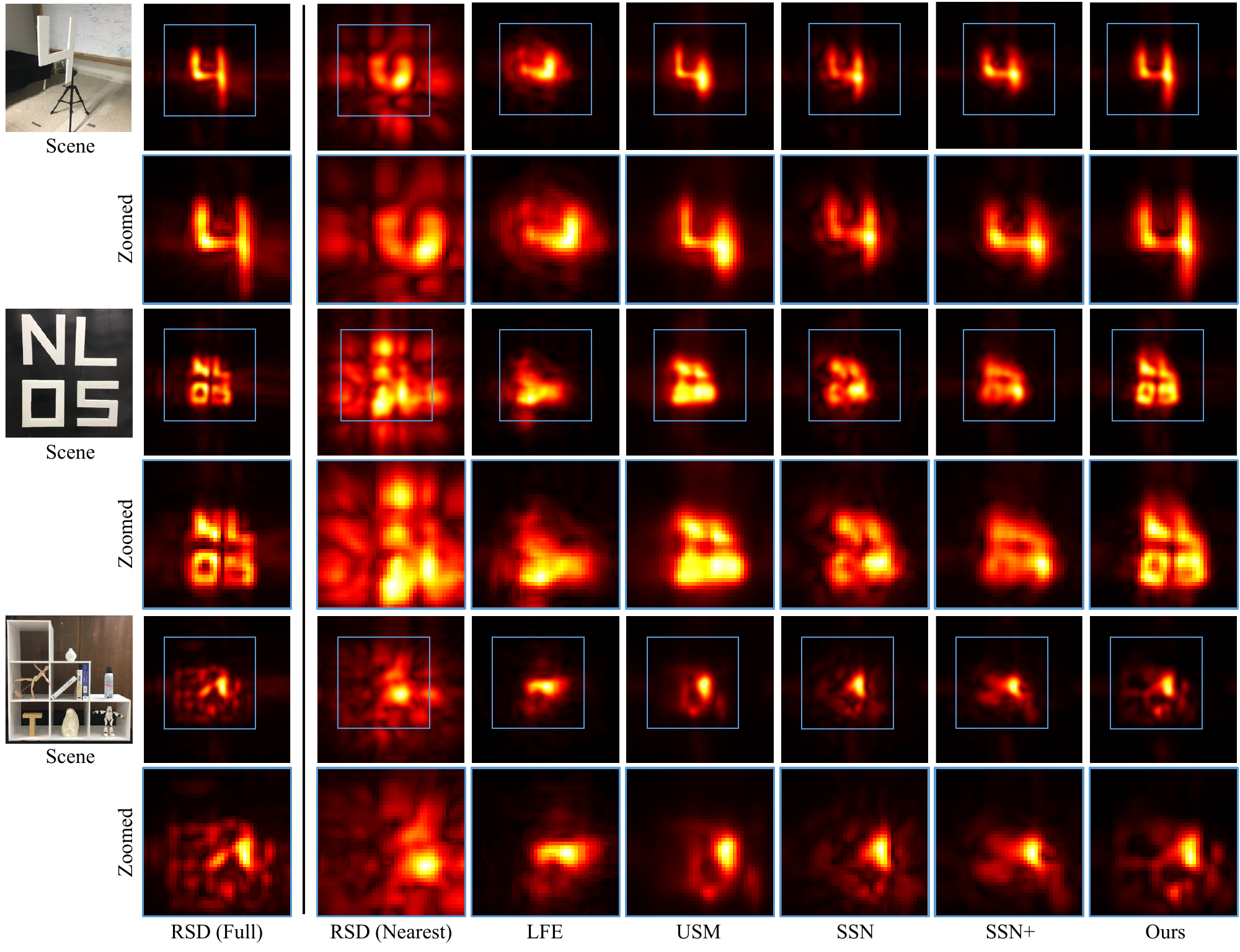}
    \end{center}
    \vspace{-10pt}
    \caption{
    Qualitative results on ``4'', ``NLOS'', Shelf instances from the non-confocal real-world dataset \cite{liu2020diffraction}.
    We deliver the results in the $8 \times 8$ sparse sampling scenario.
    Our model successfully reconstructs clean and detailed structures of the objects, while other baseline methods produce noisy outputs, often failing to reveal details of the objects.
    }
    \label{fig:supp_nonconfocal_other_instances}
    \vspace{-10pt}
\end{figure*}

%% file: Supple/Sections/additional_discussion.tex
\section{Additional Discussion}
\parabf{\textbf{Societal impact.}}
Revealing objects that are previously occluded in conventional line-of-sight systems can yield significant benefits across multiple domains, including autonomous driving and medical imaging.
For example, in the field of medical imaging, NLOS imaging has the potential to reconstruct organs that were previously obscured, thereby expanding the scope of surgical procedures and medical assessments possible.
It is crucial to emphasize that the safety issues associated with the laser utilized in NLOS imaging systems should be carefully considered.
Researchers exploring NLOS imaging must remain vigilant about potential misuses, such as violating personal privacy or terrorism.

\parabf{Supervision of the reconstructed volumes.}
While our method is supervised in the measurement space, it can be easily extended to further refine the reconstructed volumes, as the employed RSD propagator for reconstructing hidden volumes is also a linear operator. This can be accomplished by either (1) utilizing additional supervision for the reconstructed volumes after the RSD propagator, or (2) integrating the volume refinement module used in LFE \cite{chen2020lfe} after our enhancement network.
Such extensions would provide opportunities to exploit additional priors, such as the sparsity of the reconstructions or the band-limited property of the RSD convolution kernel \cite{jiang2021ring}. Although these extensions incur additional training costs, including increased training time and GPU memory consumption due to the use of FFT operators, we believe that exploring these extensions can be an interesting direction for future research.

\parabf{\textbf{NLOS imaging systems and additional sensors.}}
\label{sec:additional_sensors}

It is worth mentioning that adjusting sampling distances, the number of samplings, and scanning areas of scanning grids are readily available functionalities in NLOS imaging systems, introduced in previous works \cite{o2018lct, lindell2019fk}.
This enables our model to be highly compatible with these imaging systems, without requiring additional modification of the hardware.
In addition, the subsampling process in our evaluations faithfully replicates actual scanning procedures in partial sampling scenarios. We achieve this by completely removing certain pixels from inputs, placed between uniformly sampled pixels (strides) or placed out of target regions (cropping).
Consequently, we anticipate that our model will seamlessly integrate with previous NLOS imaging systems and address practicality issues inherent in previous scanning procedures.

\input{Supple/Partials/figure_confococal_downx}
For the confocal evaluation in the main paper, we presented the results on real-world measurements with initial $2\times$ average downsampling, which results in the increase of the exposure time per pixel.
Here, we also present results on the real-world measurements of Bike \cite{lindell2019fk} without the initial downsampling.
As shown in \Fref{fig:supp_confocal_downx}, results on the measurements without initial downsampling are almost similar to the results reported in the main paper with a similar exposure time.

In recent years, within the field of NLOS imaging, several methods have demonstrated the advantages of utilizing arrays of SPAD sensors, \eg possibility of reducing laser power and increasing detected photon counts with same scanning time.
Nevertheless, employing a large number of SPAD sensors would introduce expensive additional costs for the equipment, which would not be affordable in many real-world scenarios and applications.
Our model presents itself as an alternative yet effective solution for such cases, given the demonstrated effectiveness of our method under the single-pixel scanning setup throughout the manuscript.
Furthermore, we would like to note that our method is not bounded to a certain acquisition setup.
 The same training and inference pipeline can be seamlessly extended to a multiple-sensor configuration, with the only prerequisite being the preparation of suitable datasets using the NLOS renderer. 
We anticipate that our model can contribute to the reduction of required exposure time and the number of sensing pixels, consequently mitigating hardware costs for NLOS imaging with multiple sensors.

\parabf{\textbf{Neural networks for NLOS imaging.}}
Previous learning-based NLOS methods are primarily designed in an end-to-end manner, wherein neural networks learn to refine propagated feature volumes and directly predict 2D outputs.
These methods demonstrate high-quality results when the employed physical propagators can effectively transform spatiotemporal features in the measurement space to spatial feature volumes.
However, in partial sampling scenarios, as evidenced by the results of interpolation methods, existing inverse NLOS methods struggle to produce accurate outputs.
This subsequently leads to failures of previous end-to-end learning-based methods.

While USM \cite{li2024usm} achieves improved performance by incorporating a signal recovery network before the volume refinement module, its signal recovery network lacks sufficient feature dimensions and capacity to learn rich and noise-robust representations in the measurement space.
In contrast, rather than training end-to-end networks, we prioritize the extraction of rich, informative, and noise-robust representations in the measurement space.
This is accomplished through the denoising autoencoder scheme, and the phasor-based frequency filtering to extract informative signals from measurements.
Our model demonstrates compelling results across all evaluation scenarios, highlighting the importance of learning informative and noise-robust representations in the measurement space.

%% file: Supple/Partials/figure_confococal_downx.tex
\begin{figure}
\centering
    \includegraphics[width=0.6\linewidth]{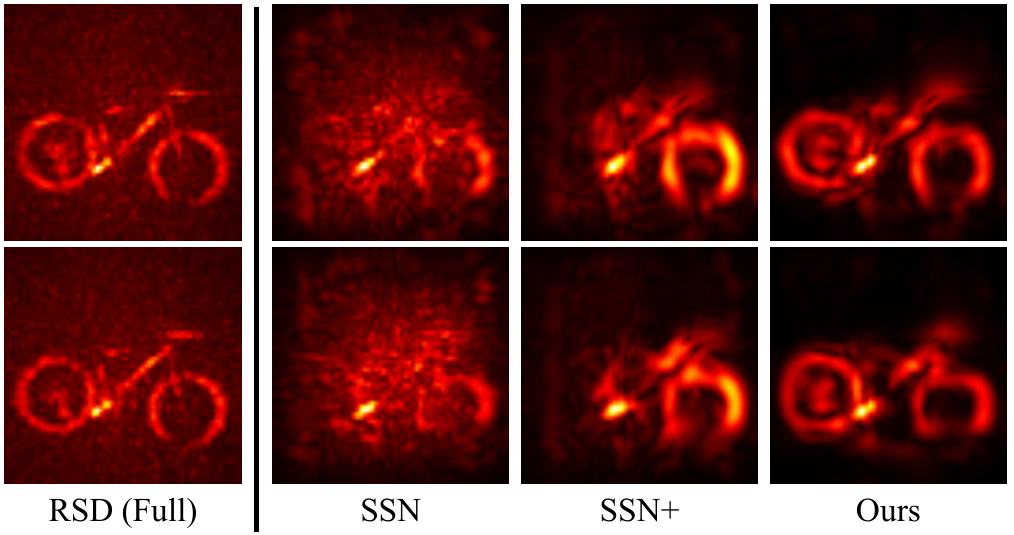}
    \caption{
    Results on the confocal real-world measurements of Bike from the Stanford dataset [21].
    \textbf{(top)} Results on measurements with initial $2\times$ downsampling as reported in the main paper (55 \textrm{ms} exposure per pixel). \textbf{(bottom)} Results without initial downsampling (41 \textrm{ms} exposure per pixel).
    All methods deliver almost similar results regardless of the initial downsampling with a similar exposure time.
    }
    \label{fig:supp_confocal_downx}
\end{figure}

%% file: Supple/Sections/error_bar.tex
\section{Error Bar}
\label{sec:error_bar}
To validate the reproducibility of our model, we report the quantitative results on the synthetic dataset under all scanning setups, with 5 different random seeds. All experiments in the main paper are conducted with a fixed random seed ``123456''. We additionally train and validate our model with 4 sequential random seeds: from seed ``1'' to ``4''. As shown in \Tref{table:supp_error_bar}, all of our models trained with 5 different random seeds report similar quantitative results, showing less than approximately $0.05$ standard deviation for PSNR, $0.0012$ for SSIM, and $0.0003$ for RMSE.
All ablation results are reported with the same fixed random seed ``123456'' without hand-picking.

\input{Supple/Partials/table_error_bar}

\clearpage
\clearpage

%% file: Supple/Partials/table_error_bar.tex
\begin{table*}[t!]
\setlength{\tabcolsep}{4pt}
\centering
\small
\caption{Quantitative results on the synthetic dataset with various random seeds. We use a fixed random seed ``123456'' throughout all experiments in the main paper and the supplementary material. We additionally deliver the results with sequential random seeds, ``1'', ``2'', ``3'', ``4'', to test the reproducibility.
}
\resizebox{\columnwidth}{!}{
\begin{tabular}{c|ccc|ccc|ccc|ccc}
\toprule
\multirow{2}{*}{Seed} & \multicolumn{3}{c|}{Conf-16} & \multicolumn{3}{c|}{Conf-8} & \multicolumn{3}{c|}{Conf-small} & \multicolumn{3}{c}{Non-16} \\
& PSNR$\uparrow$ & SSIM$\uparrow$ & RMSE$\downarrow$ & PSNR$\uparrow$ & SSIM$\uparrow$ & RMSE$\downarrow$
& PSNR$\uparrow$ & SSIM$\uparrow$ & RMSE$\downarrow$ & PSNR$\uparrow$ & SSIM$\uparrow$ & RMSE$\downarrow$ \\
\hline

 ``1'' & 32.01 & 0.8949 & 0.0894 & 28.06 & 0.8455 & 0.0962 & 28.37 & 0.8577 & 0.0968 & 37.43 & 0.9626 & 0.1414 \\
 ``2'' & 32.03 & 0.8950 & 0.0897 & 28.07 & 0.8467 & 0.0963 & 28.46 & 0.8569 & 0.0965 & 37.40 & 0.9621 & 0.1415 \\
 ``3'' & 32.03 & 0.8960 & 0.0893 & 28.09 & 0.8493 & 0.0956 & 28.40 & 0.8577 & 0.0967 & 37.41 & 0.9622 & 0.1415 \\
 ``4'' & 32.02 & 0.8963 & 0.0894 & 28.10 & 0.8475 & 0.0964 & 28.45 & 0.8578 & 0.0966 & 37.42 & 0.9622 & 0.1417 \\
 ``123456'' & 32.02 & 0.8949 & 0.0892 & 28.07 & 0.8472 & 0.0962 & 28.31 & 0.8556 & 0.0969 & 37.45 & 0.9625 & 0.1414 \\
\hline
Mean & 32.02 & 0.8954 & 0.0894 & 28.08 & 0.8472 & 0.0961 & 28.40 & 0.8571 & 0.0967 & 37.42 & 0.9623 & 0.1415 \\
Std. & 0.01 & 0.0006 & 0.0002 & 0.01 & 0.0012 & 0.0003 & 0.05 & 0.0008 & 0.0001 & 0.02 & 0.0002 & 0.0001 \\
\bottomrule

\end{tabular}
}
\label{table:supp_error_bar}
\end{table*}